\documentclass[10pt,journal,compsoc]{IEEEtran}

\usepackage{styles/extra_style}

\title{Stabilizing Spiking Neuron Training}
\date{ }

\author{Luca Herranz-Celotti\qquad Jean Rouat \\ Universit\'e de Sherbrooke, Canada \\ \{luca.celotti, jean.rouat\}@usherbrooke.ca 
}

\begin{document}

\maketitle

\begin{abstract}
Stability arguments are often used to mitigate the tendency of learning algorithms to have ever increasing activity and weights that hinder generalization. However, stability conditions can clash with the sparsity required to augment the energy efficiency of spiking neurons. Nonetheless it can also provide with solutions. In fact,
spiking Neuromorphic Computing uses binary activity to improve Artificial Intelligence energy efficiency. However, its non-smoothness requires approximate gradients, known as Surrogate Gradients (SG), to close the performance gap with Deep Learning. Several SG have been proposed in the literature, but it remains unclear how to determine the best SG for a given task and network.
Thus, we aim at theoretically define the best SG a priori, through the use of stability arguments, and reduce the need for grid search. 
In fact, we show that more complex tasks and networks need more careful choice of SG, even if overall the derivative of the fast sigmoid outperforms other SG across tasks and networks, for a wide range of learning rates.
We therefore design a stability based theoretical method to choose initialization and SG shape before training on the most common spiking architecture, the Leaky Integrate and Fire (LIF). Since our stability method suggests the use of high firing rates at initialization, which is non-standard in the neuromorphic literature, we show that high initial firing rates, combined with a sparsity encouraging loss term introduced gradually, can lead to better generalization, depending on the SG shape.
Our stability based theoretical solution, finds a SG and initialization that experimentally result in improved accuracy. We show how it can be used to reduce the need of extensive grid-search of dampening, sharpness and tail-fatness of the SG. We also show that our stability concepts can be extended to be applicable on different LIF variants, and also within the DECOLLE and fluctuations-driven initialization frameworks.
\end{abstract}

\section{Introduction}

Stability has become one of the major tenets for the understanding of learning algorithms \cite{hochreiter1991untersuchungen, bengio1994learning, hochreiter1997long, hochreiter2001gradient, glorot2010understanding, orthogonal_initialization, he2015delving, roberts2022principles}. In fact, they can easily present a tendency to have quickly increasing values for the activity, weights and gradients, that hamper their ability to perform. This is known as the representation and gradient explosion problem, and several techniques aim at stabilizing it. However, when stability notions are applied to the Spiking Networks (SNNs) used in Neuromorphic Computing \cite{henderson2020towards, blouw2019benchmarking, 9395703, lapique1907recherches,izhikevich2003simple}, a dilemma can arise. In fact these are usually utilized to take advantage of highly energy efficient devices that benefit especially from having very sparse activity, and a sparsity principle might at times seem in conflict with a stability principle when training SNNs. It is in fact often reported that spiking layers tend to go silent with depth if initialized to encourage sparse activity \cite{mcrnorm, rossbroich2022fluctuation}, which causes very sparse initializations to perform worse on deeper SNN \cite{rossbroich2022fluctuation} than they do on shallow SNN \cite{zenke2021remarkable}.
However, stability arguments can also fortuitously provide solutions to open problems in SNNs and provide theoretical justification. For example, SNN's binary activity, makes them challenging to train, as it is not differentiable, and makes well established learning methods of gradient backpropagation seemingly inapplicable \cite{robbins1951stochastic, kiefer1952stochastic,adam}.
A common solution is to introduce an approximation of the binary activity's derivative, referred to as the Surrogate Gradient (SG) \cite{bohte2002error, esser2016convolutional, zenke2018superspike, lsnn}.
While the use of SG has been widely adopted in the field of Neuromorphic Computing, there has been limited progress in establishing a theoretical foundation for its choice \cite{surrogate2019, zenke2021remarkable}. The current practice is to choose a SG empirically, with the goal of improving performance, but with no understanding of the underlying principles.
In fact, it is common practice to pick one SG for all the experiments \cite{spikingbohte,courbariaux2016binarized,lsnn,zenke2018superspike,zenke2021remarkable,yin2021accurate}, and possibly explore the effect of changing its width (sharpness) \cite{zenke2018superspike} or its height (dampening) \cite{lsnn}. It is in this context that a stability argument could be introduced to inform us a priori on what represents a good SG.
Our work proposes to justify initialization methods for spiking neurons to balance the gradients across time. With such balancing, robustness is obtained in relation with time backpropagation.

To ensure that information is balanced through time during training, several hyper-parameters must be carefully tuned at initialization, including the SG shape. This is why an initialization method influences the choice of the best SG, which becomes apparent in our theoretical development. 
On the other hand, stability can also come at the cost of reduced reactiveness to important new stimuli. Our proposed method seeks to achieve stability while also fostering reactiveness. Moreover, we introduce an unconventional approach by initializing the network with high firing rates, 
not commonly seen in the neuromorphic literature. We leverage the fact that SG curves typically reach their maximum value when the neuron fires, so by keeping the voltage close to the firing threshold, we achieve stronger gradients. Furthermore, we show that  training with an additional sparsity encouraging loss term, can lead to the desired high sparsity on the test set, despite an initial low sparsity, while improving  generalization. This hints at the possibility of a total benefit, in terms of both performance and energy efficiency at test recall.


Our contributions are therefore:
\begin{itemize}
    \item We show that the choice of SG becomes increasingly important as task and network complexity increase;
    \item We observe that the derivative of the fast sigmoid is a resilient SG;
    \item We show that high  initialization firing rates can improve generalization with low test firing rates;
    \item We provide stability-based constraints on the LIF weights and SG shape that improve final performance;
    \item Our stability-based theory predicts optimal SG features on the LIF network;
    \item We show how such reasoning can be extended to cover different spiking neuron definitions and lead to improved generalization.
\end{itemize}

\begin{figure}
    {\footnotesize \hspace{-2.5cm}(a)\hspace{3.cm}(b)
    }
    \centering
    \includegraphics[width=.5\textwidth]{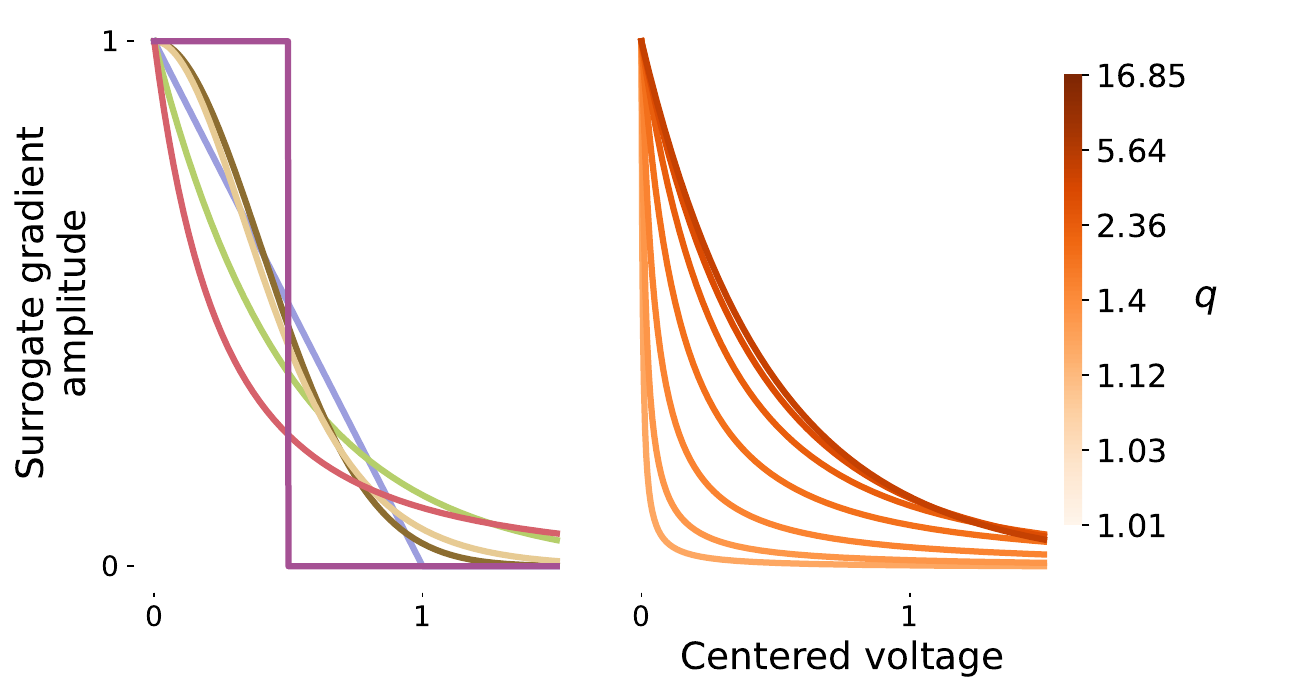}
    \includegraphics[width=.47\textwidth]{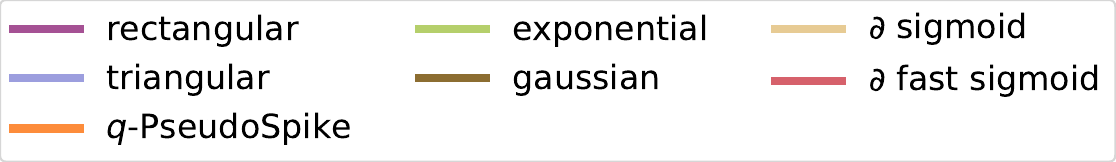}

    \caption{\textbf{Surrogate Gradient shapes.} To stabilize a network we have to stabilize also the backward pass. However the LIF, as a spiking neuron, has an undefined backward pass and we need Surrogate Gradients (SG) to approximate it. Panel (a) shows the SG investigated in this work, and (b) the tail dependence of our $q$-PseudoSpike SG for $q\in[1.01, 16.85]$. The SG considered are symmetrical around $v_t=y_t-\vartheta=0$, so we only plot half the curve (centered voltage $v_t>0$). 
    }
    \label{fig:methodo}
\end{figure}

\section{Methods}

\subsection{Representation and Gradient Stability}
The earliest form of the stability analysis in the study of neural networks is often attributed to \cite{hochreiter1991untersuchungen} who identified that the tendency of gradients to compose exponentially with depth and time was behind the difficulty in training classical recurrent neural networks. For this reason this problem is often named the Exploding and Vanishing Gradient Problem (EVGP). However this name does not emphasize the need to stabilize the forward pass, and the intermediate tensors often referred to as representations. This is probably because infinitesimally, the representations are well described by the gradient, the backward pass, which is a fair justification for fully differentiable neuron models. However, for non-differentiable neuron models, the connection between forward and backward pass is not as simple to make.
Probably the most well known results of this line of research are the Glorot and He initializations ~\cite{glorot2010understanding, he2015delving}, for linear and ReLU fully-connected feed-forward networks, who provide the exact mean and variance that the connection matrices need to have at initialization to avoid the exponential composition of the gradients. By requiring the mean of the representations to remain equal to zero and variances equal to one per layer, the gradients are guaranteed not to compose exponentially with depth. 

However, this type of analysis has never been done before for spiking neural networks. Instead, theoretical justification for recurrent networks initialization has been proposed for the LSTM \cite{lstm_initialization}, and other non spiking recurrent networks \cite{hochreiter2001gradient, arjovsky2016unitary, pascanu2013difficulty}. 
In practice,  \cite{zenke2021remarkable} samples a $Var[W_l]=1/3n_{l-1}$ Uniform, while \cite{lsnn} a $Var[W_l]=1/n_{l-1}$ Normal distribution, for similar spiking models.
Only recently a similar method with emphasis on the sparsity of activity has been developed by \cite{rossbroich2022fluctuation} to initialize the connection matrices of spiking neural networks. However, to our knowledge, this approach has not been used to determine the SG shape.

\subsection{Surrogate Gradients for Spiking Neurons}
\noindent\textbf{Surrogate Gradients.}
A problem that comes by using spiking neural networks is that they are not differentiable, so, training with gradient descent would not be possible without a patch to fix it. One patch is to use approximate gradients, often referred to as Surrogate Gradients (SG).
The non-differentiability appears in spiking networks because a spike is produced when the voltage surpasses the threshold, which mathematically is often described through a Heaviside function, $\tilde{H}(v)$, that is  zero for $v<0$ and one for $v\geq 0$. We use the tilde to remind that a SG is used for training, defined as $d\tilde{H}(v)/dv = \gamma f(\beta\cdot v)$, where $\beta$ is the sharpness, $\gamma$ the dampening, $f$ is the shape of choice and $\cdot$ the scalar product. Thus, $\gamma$ controls the amplitude of the SG, and $\beta$ controls the width and, unless explicitly stated, we set both to one. Instead, $f$ is usually defined as maximal at $f(0)$ and smoothly decaying to zero to infinity. As a result, a high sharpness, mostly passes the gradient for $v$ close to zero, while low sharpness also passes the gradient for a wider range of voltages. Importantly, the SG still allows to pass the gradient when the neuron has not fired. 

The SG shapes $f$ we investigate are (1) rectangular \cite{hubara2016binarized}, (2) triangular \cite{esser2016convolutional, lsnn}, (3) exponential \cite{Shrestha2018SLAYERSL}, (4) gaussian \cite{taulsnn}, (5) the derivative of a sigmoid \cite{zenke2021remarkable}, and (6) the derivative of a fast-sigmoid, also known as SuperSpike \cite{zenke2018superspike}. These are the most popular choices, and other curves like the arctan, are expected to behave similarly to the smooth alternatives, such as the gaussian or SuperSpike. Their curves are plotted in Fig. \ref{fig:methodo} and their equations can be found in App.~\ref{app:surrogate}. To make the comparison between different SG more clear, $f$ is chosen to have a maximal value of $1$ and an area under the curve of $1$. We also propose a generalization of the derivative of the fast-sigmoid, that we call $q$-PseudoSpike SG. Its tail fatness is controlled by a hyper-parameter $q$ and we use it to study tail dependence in section \ref{sec:heavy_tails}. Notice that computing an exponential has a time complexity upper bound of $O(\mu(n) \log n)$, with $\mu(n)$ being the time complexity upper bound of n-bit integer multiplication \cite{brent1976fast, expcomplexity}, for a relative error $<O(2^{-n})$. Despite (3, 4, 5) having an exponential in their definition,  experimentally we did not find any difference in speed with the rest and therefore the computation of the exponential did not represent a speed bottleneck compared to other operations in the architecture.
We use those SG to train variants of the most common spiking neuron, the Leaky Integrate and Fire (LIF) neuron.

\noindent\textbf{Leaky Integrate and Fire spiking neuron.}
Arguably the simplest spiking neuron is the LIF \cite{lapique1907recherches, gerstner2014neuronal, wozniak2020deep}. It is defined as $\boldsymbol{y}_t = \boldsymbol{\alpha}_{decay} \boldsymbol{y}_{t-1}(1-\boldsymbol{x}_{t-1}) + \boldsymbol{i}_{t-1} $ where $\boldsymbol{i}_{t}=W_{rec}\boldsymbol{x}_{t} + W_{in}\boldsymbol{z}_t + \boldsymbol{b}$, and $y_t$ is the neuron membrane voltage, using \cite{glorot2010understanding, he2015delving} notation. We define $x_{t} = \sigma(y_{t})= \tilde{H}(y_{t}-\vartheta)= \tilde{H}(v_{t})$ as the spiking activity, where  $\vartheta$ is the spiking threshold, $v_t=y_{t}-\vartheta$ the centered voltage, and $\tilde{H}(v_{t})$ a Heaviside function with SG. The term $(1-\boldsymbol{x}_{t-1})$ represents a hard reset, that takes the voltage to zero after firing. The input $z_t$ can represent the data, or a layer below. It is common to write $\alpha_{decay}= 1-\frac{dt}{\tau_m}$, where $dt$ is the computation time, $\tau_m$ the membrane time constant, and to multiply the other terms by biologically meaningful constants, that we compress for cleanliness. Each neuron can have its own speed $\alpha_{decay}$, intrinsic current $b$ and $\vartheta$. In this work, all the parameters in the LIF definition are learnable.

We denote vectors as $\boldsymbol{a}$, matrices as $A$, and their elements as $a$.  In a stack of $L$ layers, we add an index $l$ to each parameter and variable. The matrix $W_{rec,l}\in\mathbb{R}^{n_{l}\times n_{l}} $ connects neurons in the same layer, with zero diagonal, and $W_{in,l}\in\R^{n_{l}\times n_{l-1}} $ connects the layer with the layer below, or the data if $l=0$, where $n_l$ is the number of neurons in layer $l$. We use curved brackets $A(\cdot)$ for functions, and square brackets $A[\cdot]$ for functionals that depend on a probability distribution. We use interchangeably $\overline{a} = Mean[a]$, $\hat{a}=Max[a]$, and $\check{a}=Min[a]$ for any variable $a$. 
Since the equation only depends on the previous time-step and layer, the probability distribution is a Markov chain in time and depth. Therefore the statistics we discuss are computed element-wise with respect to the distribution $p(y_{t,l}|t,l) = p( \boldsymbol{y}_{t-1, l},\boldsymbol{z}_{t,l-1},W_{rec,l}, W_{in,l},b_l, \alpha_{decay,l},\vartheta_l |t,l)$.

Throughout the article we use $\rho$ to be the average activity of a layer, and therefore its firing rate, as the mean across the time, neurons and minibatch samples. Mathematically, it is simply the mean of the spiking activity: $\rho=\overline{x}$, and therefore we consider it to be unitless, as a firing probability. In the neuromorphic literature it is common to assume \mbox{$dt=1ms$}, and each time-step to have that duration, which makes a \mbox{$\rho=1/2$} equivalent to $500Hz$. Hence the sparsity could be quantified as $1-\rho$, since lower firing rate, means sparser activity. We stress that we are not looking for biologically plausible values of $\rho$, and we are instead concerned by the computational and learning capabilities of the system. 

Additionally, to emphasize the need for our line of work, we show how the impact of the SG choice becomes more unpredictable as we increase the task complexity and the network complexity.
To do that, we briefly make use of two extra neuron definitions. When a LIF is upgraded with a dynamical threshold to maintain longer memories, we have the Adaptive LIF (ALIF) \cite{gerstner2014neuronal, lsnn}. Moreover, we propose the spiking LSTM (sLSTM), defined by changing the LSTM \cite{hochreiter1997long} activations by neuromorphic counterparts. The equations for both the ALIF and the sLSTM can be found in App.~\ref{app:alifsLSTM}. We quantify their complexity as in \cite{yin2021accurate}, and Tab.~\ref{tab:complexities}, by the number of operations performed per layer. However, we apply our stability method only to the LIF.

\subsection{Surrogate Gradient Stability}
When studying the stability of representations and gradients, it is typical to study the dependence of the mean and the variance of representations and gradients with depth and time, to look for initialization hyper-parameters that eliminate such dependence, and avoid an exponential explosion or vanishing altogether. Typically both forward and backward pass result in a similar constraint on the mean and variance of the connection matrix~\cite{glorot2010understanding, he2015delving}. In this work we study mean and variance of the representations with conditions I and II and maximum value and variance of the gradients with conditions III and IV, all of them on the LIF network. Conditions I and II result in constraints on the connection matrices, while conditions III and IV result in constraints on the SG shape.
In a sense we are interested in stabilizing as many quantities susceptible to exponential composition as possible, being the mean and the variance the most typical magnitudes to stabilize. In fact, we will see experimentally if applying all the conditions outperforms applying any single one of them.
We present the mathematical equivalent in each subsection, and the derivation details in the Appendix, but in summary
\begin{enumerate}[label=\Roman*]
\itemsep0em 
    \item The voltage should hit the most sensitive part of the SG;
    \item Recurrent and input variances should match;
    \item Gradients must have equal maxima across time;
    \item Gradients must have equal variance across time.
\end{enumerate}

Given that these conditions result in specific values to assign at initialization, they do not imply additional training complexity. The method guides the search for hyperparameters, which is something that has to be done in any case when implementing a network.

\noindent\textbf{Condition I: Recurrent matrix mean sets the firing rate.}
In the gradient learning literature, it is standard to choose initializations that place the pre-activation activity 
to hit the most sensitive part of the activation, which usually is around zero 
~\cite{glorot2010understanding, he2015delving, roberts2022principles, Hanin2018HowTS, ioffe2015batch, ba2016layer}.
Moreover, notice that SG curves reach their highest when the neuron fires, Fig.~\ref{fig:methodo}. Thus, if the voltage stays close to firing, the gradient is stronger. This is always so if the centered voltage satisfies $Median[v]=0$ and $Var[v]=0$. However, $Var[v]=0$  turns off all higher moments, thus, we only assume $Median[v]=0$ as the mathematical equivalent of our desiderata. When (I) is applied to a LIF network (see App.~\ref{app:means}), the mean of the recurrent weight matrix fixes $\rho_i$, further assuming $\overline{w}_{in}=0$,  $\boldsymbol{b}=0$, the approximation $Mean[v]\approx Median[v]$, and constant $\overline{i}_t$ over time, we find
{\small
\begin{align*}
    &\overline{w}_{rec}=
    \frac{1}{n_{l}-1}\Big(2-\alpha_{decay}\Big)\vartheta && \text{(I)}
\end{align*}
}
The assumption $Mean[v]\approx Median[v]$, can be justified by noticing that if $v$ is sampled from a unimodal distribution with the first two moments defined, then $|Mean[v]-Median[v]|\leq\sqrt{0.6Var[v]}$ is true \cite{basu1997mean}. Experimentally, we observe always unimodal distributions on the datasets we define in section \ref{sec:datasets}, that verify $|Mean[v]-Median[v]|\leq\sqrt{c\Var[v]}$, with $c=10^{-4}$ for the SHD task, $c=3\times10^{-2}$ for the sl-MNIST task, and $c=10^{-3}$ for the PTB task, with and without (I), much closer than only assuming the unimodality.

This choice of initialization results in firing rates of \mbox{$\rho=1/2$}, or 500Hz for the common assumption of $1ms$ per time step, which are considered high in the neuromorphic literature, that has a strong preference for lower firing rates. However, it is common to try to prevent the spiking neurons from turning too sparse or completely off through a loss term \cite{lsnn, zenke2021remarkable}, especially given that spiking layers tend to go silent with depth \cite{mcrnorm, rossbroich2022fluctuation}, which causes very sparse initializations to perform worse in deeper SNN \cite{rossbroich2022fluctuation}. Also, deep SNN are observed to perform worse with sparsity than their shallow counterparts, both at initialization \cite{rossbroich2022fluctuation} and after training \cite{zenke2021remarkable}. Some have utilized $\rho=1/2$ at initialization without explicitly acknowledging its use on feed-forward LIF with local losses, such as in DECOLLE \cite{decolle}. We stress that high sparsity on the test sets can be achieved, after training, even if we start with high firing rates at initialization, before training. Therefore, we study two settings: starting with a variety of firing rates at initialization $\rho_i$, we proceed to train with and without a Sparsity Encouraging Loss Term (SELT).
The SELT is a mean squared error between a target firing rate $\rho_t=0.01$ and the layer firing rate $\rho_l=\overline{x}_{t,l}$, such that $\mathcal{L}_{SELT} = \lambda/L\sum_l(\rho_l-\rho_t)^2$, where $\lambda$ is a multiplicative constant. 
To achieve different $\rho_i$, we pre-train $\boldsymbol{b}_l$ on the dataset of interest, holding the other parameters untrained, using only the SELT without the classification loss. 
The coefficient $\lambda$ to multiply the loss term is chosen to make all losses comparable only when the task is learned, to let the network focus first on the task and then on the sparsity. We therefore choose as the multiplicative factor the minimal training loss achieved without SELT, since the SELT takes values between zero and one. We switch on the SELT gradually during training. The switch starts as zero, and moves linearly to one between $1/5$ and $3/5$ of training, and stays on thereafter. We focus on the $\partial$ fast-sigmoid and the SHD task in the main text, but we show different SG and tasks in App.~\ref{app:more_sparsity}.
Finally, we measure the Pearson correlation of the firing rate before and after training ($\rho_i, \rho_f$ for initial and final) with loss after training, on the test set. We therefore use this study to understand the impact of the use of high firing rates at initialization, as suggested by condition I, despite not being common practice in the neuromorphic literature.

\noindent\textbf{Condition II: Recurrent matrix variance can  make recurrent and input contribution to voltage comparable.}
Applying the Glorot and He method would suggest to set the variance of each layer output to one, to avoid exponential composition. However, the output of the forward pass of a spiking neuron is always strictly one or zero, so, it cannot compose exponentially. Instead, we propose to set the variance of the input to remain similar to the recurrent contribution to the variance. This can be understood as the maximally non-informative Bayesian prior decision to make, when the nature of the task at hand is not known. We describe it mathematically as $Var[W_{rec}x_{t-1}] = Var[W_{in}z_t]$. In a LIF network, we show in App.~\ref{app:ineqrec}, further assuming $\rho_l =1/2$,  $\overline{w}_{in}=0$, and computing $Var[z_t]$ and $\overline{z}_t$ on the training set, that this statement conduces to
{\small
\begin{align*}
    &Var[w_{rec}] =  2(Var[z_t] + \overline{z}_t^2)\frac{n_{l-1}}{n_{l}-1}Var[w_{in}] - \frac{1}{2}\overline{w}_{rec}^2 && \text{(II)}
\end{align*}
}
Therefore, condition II can be  used to set the variance of the recurrent matrix to makes both, input and recurrent contributions equal.

\noindent\textbf{Condition III and IV: Dampening and sharpness set gradient maximum and variance.}
We have so far described how to stabilize the forward pass.
Instead, to control the backward pass, we look for constraints to obtain stable gradients with time. We describe mathematically (III) as $Max[\frac{\partial}{\partial \theta}y_t] = Max[\frac{\partial}{\partial \theta}y_{t-1}]$ and (IV) as $Var[\frac{\partial}{\partial \theta}y_t] = Var[\frac{\partial}{\partial \theta}y_{t-1}]$. On a LIF network, they set the dampening and the second moment of the SG that keep the maximum and variance of the gradient stable with time (App.~\ref{app:sgmax}, \ref{app:backward}). Sharpness and tail-fatness are linked to the SG second moment (App.~\ref{app:ivexp}). Assuming $\sigma'$ and $\frac{\partial}{\partial \theta}y_{t-1}$ as independent, and zero mean gradients at initialization, we find
{\small
\begin{align*}
    &\gamma=   \frac{1}{(n_{l}-1)\hat{w}_{rec}}\Big( 1 -\alpha_{decay} \Big)&& \text{(III)}\\
    &\overline{\sigma^{\prime 2}} = \frac{1-\frac{1}{2}\alpha_{decay}^2}{(n_{l}-1)\overline{w^2}_{rec}}  && \text{(IV)}
\end{align*}
}
To be able to clearly relate the sharpness $\beta$ with the non centered second moment of the SG, $\overline{\sigma^{\prime2}}$, we show in App. \ref{app:ivexp}, that assuming a uniform $v$ distribution gives
\begin{align}
    \label{eq:2momsigma}
    \overline{\sigma^{\prime2}}
    =&\frac{\gamma^2}{\beta(y_{max}-y_{min})}\int^{\beta(y_{max}-\vartheta)}_{\beta(y_{min}-\vartheta)} f(v)^2dv
\end{align}
Therefore, different SG  shapes $f$, will require different sharpness $\beta$ to meet condition (IV), since the result of the integration will depend on $f$. We use both equations for (III) and (IV) to make the respective predictions for the theoretically justified dampening, sharpness and tail-fatness in Fig.~\ref{fig:conditions}. For the dampening the method is simply applying the equation. For the sharpness prediction, we use the exponential SG, and as it can be seen in equation \ref{eqIVexp}, the dependence on the integration limits makes it impossible to isolate the sharpness analytically. Instead we retrieve it by finding the root of the resulting equation through gradient descent. Similarly, to find the theoretically justified tail-fatness, we use the $q$-PseudoSpike to arrive at equation \ref{eq:IVtail}, and isolate the $q$ predicted finding the root of the resulting equation through gradient descent.

\subsection{Datasets}
\label{sec:datasets}

In this work we use three tasks, that we present in increasing number of classes, which we will use as a proxy for task complexity. 
More details on the datasets can be found in App.~\ref{app:trainingdetails}.

\noindent\textbf{Spike Latency MNIST (sl-MNIST):} the MNIST digits \cite{mnist} pixels (10 classes) are rescaled between zero and one, presented as a flat vector, and each vector value $x$ is transformed into a spike timing using the transformation $T(x) = \tau_{eff}\log(\frac{x}{x-\vartheta})$ for $x>\vartheta$ and  $T(x) = \infty$ otherwise, with $\vartheta=0.2, \tau_{eff}=50 \text{ms}$ \cite{zenke2021remarkable}. The network input is a sequence of $50$ms, $784$ channels ($28\times 28$), with one spike per row.

\noindent\textbf{Spiking Heidelberg Digits (SHD):} is based on the Heidelberg Digits (HD)  audio dataset \cite{cramer2020heidelberg} which comprises 20 classes of spoken digits, from zero to nine, in English and German, spoken by 12 individuals.  These audio signals are encoded into spikes through an artificial model of the inner ear and parts of the ascending auditory pathway.

\noindent\textbf{PennTreeBank (PTB):} is a language modelling task. The PennTreeBank dataset \cite{ptb}, is a large corpus of American English texts. We perform next time-step prediction at the word level. The vocabulary consists of 10K words, which we consider as 10K classes. The one hot encoding of words can be seen as a spiking representation, even if it is the standard representation in the non neuromorphic literature.

\subsection{Training Details}

Our networks comprise two recurrent layers. The output of each feeds the following, and the last one feeds a linear readout. Our LIF network has 128 neurons per layer on the sl-MNIST task, 256 on SHD, and one layer of 1700 and another of 300 on PTB, as in \cite{wozniak2020deep}. On the complexity study with the SHD task, the ALIF has 256 neurons and the sLSTM 85, to keep a comparable number of 350K parameters. We train on the crossentropy loss.  The optimizer had a strong effect, where Stochastic Gradient Descent \cite{robbins1951stochastic, kiefer1952stochastic} was often not able to learn, and AdaM \cite{adam} performed worse than AdaBelief \cite{zhuang2020adabelief}. AdaBelief hyper-parameters are set to default, as in \cite{radford2018improving, zenke2021remarkable}. The remaining hyper-parameters are reported in App.~\ref{app:trainingdetails}. Unless explicitly stated, we use Glorot Uniform initialization. Each experiment is run 4 times and we report mean and standard deviation. Experiments are run in single Tesla V100 NVIDIA GPUs. We call our metric the mode accuracy: the network predicts the target at every timestep, and the chosen class is  the one that fired the most for the longest.




\begin{figure*}
    \centering
    \includegraphics[width=1.\textwidth]{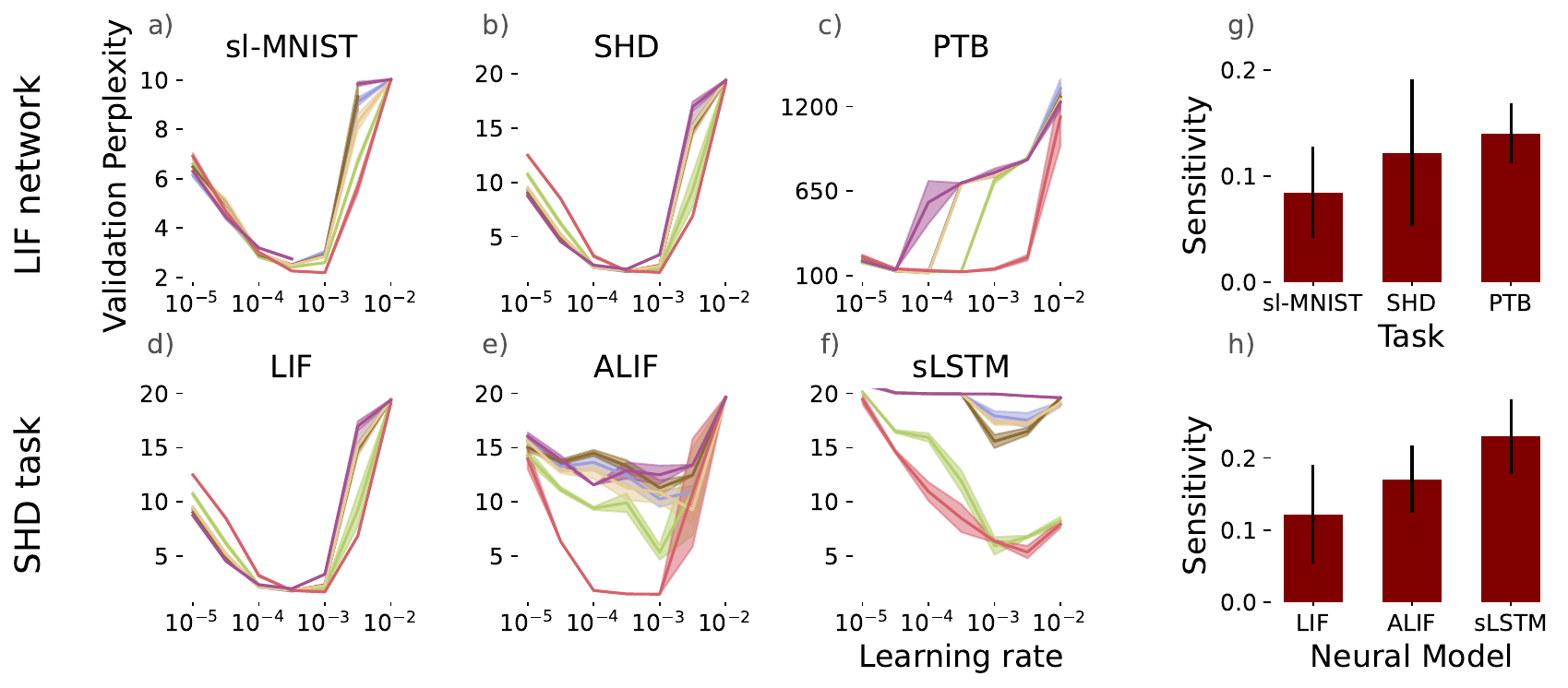}
    \includegraphics[width=1.\textwidth]{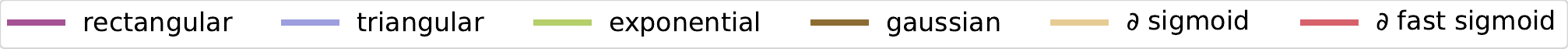}
    
    \caption{\textbf{The choice of SG becomes increasingly important as task and network complexity increase.} In order to clearly showcase the problem addressed by our work, and to understand the difficulties brought by SG training, we want to see the impact of training with different SG shapes, and how task and network complexity affect it. This will stress the need for clever rules to apply at initialization to prevent worst case scenarios. Tasks and networks are presented from left to right in order of increasing complexity, where number of classes is used as a proxy of task complexity, and the number of operations is used to quantify network complexity. We perform a grid search over SG shapes, learning rates, tasks and networks. We report lowest validation perplexity after converged training, where perplexity is a loss, so, the lower the better. Panels a-f) show  perplexity (y-axis), against learning rate (x-axis). In a-c) we fix the LIF network and change task, while in d-f) we fix the SHD task and change network. Plots b) and d) are repeated for clarity. Panels g-h) show SG sensitivity (y-axis) against task and neural model (x-axis), where we defined sensitivity in Sec. \ref{sec:sensitivity}, and it is essentially the variance of the perplexity, across SG shapes and learning rates.  a-f) Our results demonstrate that even if different SG shapes tend to agree on the optimal learning rate, the final performance can vary substantially, depending on the SG selection. Specifically, the $\partial$ fast-sigmoid seems the most resilient to changes in the learning rate, as shown in \cite{zenke2021remarkable}. g-h) Moreover, we observe that the more complex the task or the network, the higher the performance variability we see across SG shapes and learning rates. 
    Our stability criterion provides a method to carefully select SG shapes at initialization and address this issue, promoting better performance and generalization.
    }
    \label{fig:task_net_dependence}
\end{figure*}

\section{Results}

\subsection{Sensitivity increases with Complexity}
\label{sec:sensitivity}

In order to stress the difficulty of choosing the right SG, we investigate how performance changes with SG as we increase task and network complexity. We estimate the task complexity by the number of classes. Thus, if $C_T(\cdot)$ measures task complexity, $C_T(sl\text{-}MNIST)<C_T(SHD)<C_T(PTB)$. We quantify neural complexity as in \cite{yin2021accurate}, and Tab.~\ref{tab:complexities}, by the number of operations performed per layer. In essence, if $C_M(\cdot)$ measures model complexity, then $C_M(LIF)<C_M(ALIF)<C_M(sLSTM)$. To have comparable losses across tasks and networks, we normalize their validation values between 0 and 1. For that, we remove the lowest loss achieved by a network in a task for any seed and learning rate, and divide by the distance between the highest and lowest loss. We call the result the \textit{post-training normalized loss}. We call \textit{sensitivity} the standard deviation of the \textit{post-training normalized perplexity} across SG, for each learning rate. We report mean and standard deviation across learning rates.

We see in Fig.~\ref{fig:task_net_dependence}, that task and network complexity have a measurable effect on the sensitivity of training to the SG choice.  We run a grid search over learning rates and SG shapes. The sensitivity to the task is shown in the upper panels, for the LIF network. We see that different SG agree on the optimal learning rate. We also see that the $\partial$ fast-sigmoid performs well for a wider range of learning rates. The rectangular SG is competitive on some tasks, but fails to learn with most learning rates on PTB. Then we focus on network sensitivity, fixing the SHD task, lower panels. The triangular SG performs similarly to the exponential on the LIF network, while it underperforms on ALIF, and fails on sLSTM. The exponential SG matches the best SG on both the LIF and the sLSTM, but not on the ALIF. All this manifests a strong sensitivity to the SG choice. 
Surprisingly, the sLSTM lags behind the LIF and ALIF, with a comparable number of parameters. The gating mechanism devised to keep the LSTM representations from exploding exponentially, are not relevant anymore for a Heaviside that cannot explode exponentially, and might have become a computational burden. 
Incidentally, we reached spiking state-of-the-art on the PTB task with the triangular SG. Best average over 12 seeds had $122.8 \pm 10.7$ validation and $114.2 \pm 9.2$ test perplexity, and best seed had $117.2$ validation and $109.5$ test perplexity. Previous spiking SOTA on PTB was 137.7 test perplexity~\cite{wozniak2020deep}.  Fig.~\ref{fig:task_net_dependence}, g-h), confirm that there is a correlation between task and network complexity, and SG sensitivity. This stresses the importance of finding the correct SG to achieve maximal performance.

\subsection{High  initialization firing rates can improve generalization with low test firing rates}
\label{sec:ressparse}

\begin{figure}
    \includegraphics[width=.5\textwidth]{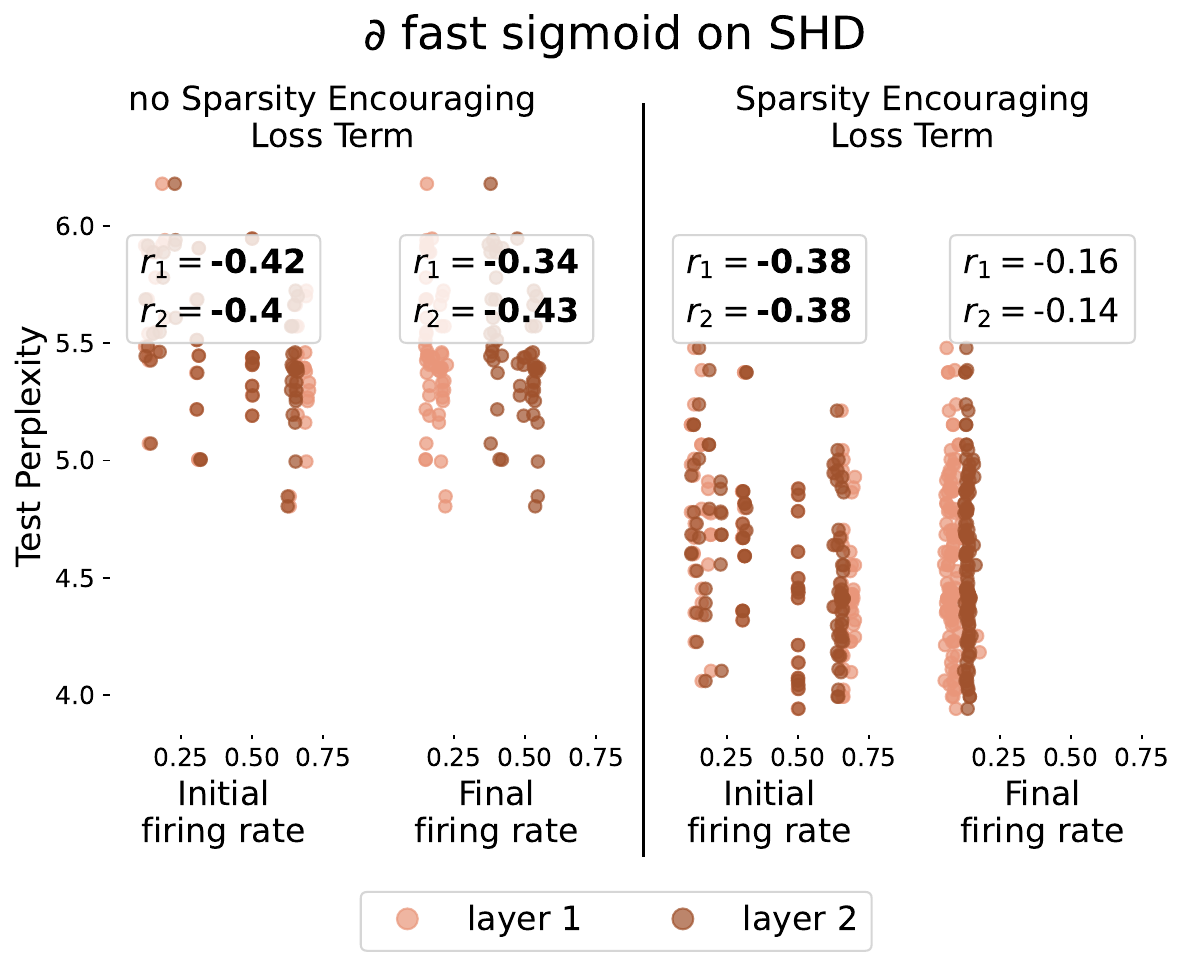}
        \caption{\textbf{High  initialization firing rates can improve generalization with low test firing rates.} Our initialization method suggests to set a high firing rate at the beginning of training, which is uncommon in the neuromorphic literature. We study if it is possible to reconcile high initialization firing rates with low firing rates on the test set. We use the SHD task and the $\partial$~fast-sigmoid SG, and measure the correlation $r_l$ of each layer $l$ firing rate with perplexity after training, on the test set. Bold correlation means $p$-value $ \leq 0.05$. On the $y$-axis we report perplexity after training on the test set, and on the $x$-axis we report initialization firing rate $\rho_i$, or final firing rate $\rho_f$, meaning the firing rate after training, also evaluated on the test set. On the two left panels, learning starts from different $\rho_i$ without a Sparsity Encouraging Loss Term (SELT), while on the two right panels a target sparsity is encouraged. In both cases, the initial firing rate correlates with final performance, and a low $\rho_f$ is achieved successfully using a SELT. Notice as well that the combination of high initial firing rate and sparsity encouragement resulted in better test loss than on the two panels on the left, suggesting that both factors acted synergistically as a regularization mechanism. We conclude that high initialization firing rates are not necessarily at odds with having sparse activity after training.}
    \label{fig:sparsity}
\end{figure}

In order to propose our stability-based theoretical method for SG choice, we want to make sure that high initial firing rates are not pernitious neither for learning nor for final sparsity. This is so, because in the neuromorphic literature training success is judged by (1) training performance and (2) activity sparsity. 
We can see in Fig.~\ref{fig:sparsity} that with and without a SELT, higher $\rho_i$ correlates with performance. 
In fact at each layer $l$, the correlation $r_l$ of the firing rate with the loss is markedly negative, and statistically significant, where we show in bold whenever $p$-value $\leq0.05$.
Notice that SELT achieved worse final train loss (not shown). However, the high $\rho_i$ combined with SELT resulted in better test loss, thus, better generalization. However, this is not consistent across SG shapes, Fig.~\ref{fig:sparsity_and_shape}, but is consistent across tasks, Fig.~\ref{fig:sparsity_and_task} App.~\ref{app:more_sparsity}. In fact, the triangular SG prefers low $\rho_i$ and the exponential SG does not show a clear trend. Incidentally, the lower layer always reaches higher sparsity, across seeds (Fig.~\ref{fig:sparsity}), SG shapes (Fig.~\ref{fig:sparsity_and_shape}) and tasks (Fig.~\ref{fig:sparsity_and_task}).

\subsection{Our stability-based constraints on the LIF weights and SG shape improve final performance.}
\label{sec:conditions}

\begin{figure}
    \includegraphics[width=.45\textwidth]{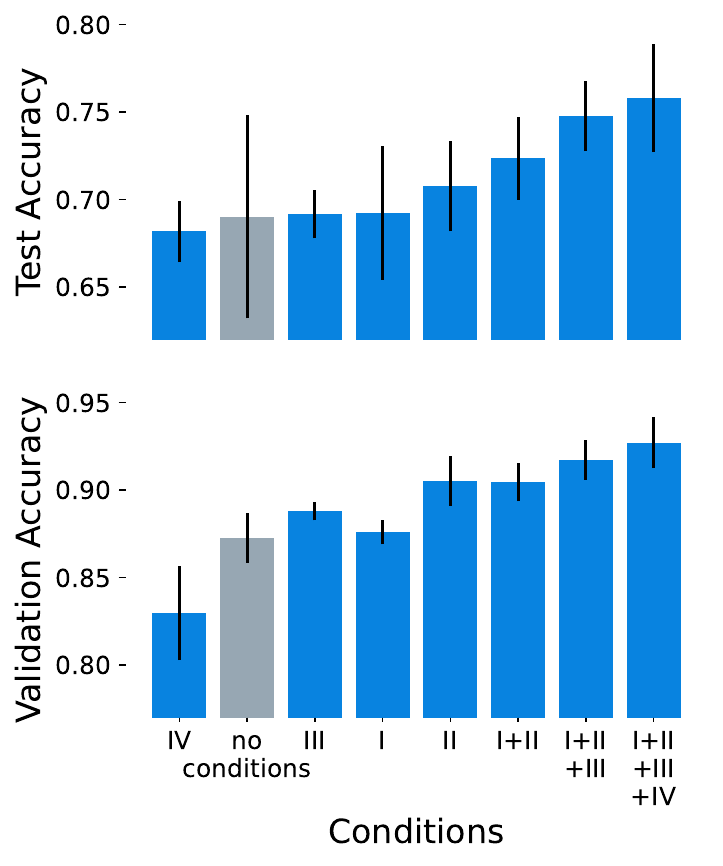}
    \caption{\textbf{Our stability-focused constraints on the LIF weights and SG shape improve  final performance.} 
    This figure illustrates our novel method for selecting SG in a Leaky Integrate-and-Fire (LIF) network to improve its stability and performance. 
    We design 4 conditions to stabilize forward and backward pass of a LIF network.~(I) requires voltages that promote higher SG values, (II) balances input and recurrent contribution to the voltage, while (III) and (IV) constrain gradient maxima and variance over time. 
    We demonstrate the effectiveness of our method for the LIF network on the SHD task, with an exponential SG and Glorot Uniform initialization, and compare training under the four stability conditions with the baseline without any conditions (shown in gray).
    Lower and upper panels show validation and test accuracies. Our results show that while condition (II) has the most significant impact on its own, all four conditions combined lead to the best performance.
    These findings suggest that our theory of LIF stabilit can reduce the need for extensive hyper-parameter search and improve the experimental performance of LIF networks.}
    \label{fig:conditions}
\end{figure}

Keeping in mind that we can exploit a low initial sparsity as a regularization mechanism, we have proposed a method for stabilizing LIF networks inspired by FFN initializations \cite{glorot2010understanding, he2015delving}, that determines initialization weights and SG shape. The four conditions we propose, result in a SG that depends on the network and the task.
Fig.~\ref{fig:conditions} shows training results with our conditions for the LIF network on the SHD task, with exponential SG, against the unconditioned baseline. Condition II improves accuracy the most when applied on its own, but the best performance is achieved with all conditions together. When all conditions are applied, a LIF network achieves a $92.7\pm1.5$ validation and $75.8\pm3.1$ test accuracy, compared to $87.3\pm1.4$ validation and $69.0\pm5.8$ test accuracy without conditions.

\begin{figure*}
    {\footnotesize \hspace{1.2cm}(a)\hspace{5.6cm}(b)\hspace{5.6cm}(c)}
    
    \centering
    \includegraphics[width=.32\textwidth]{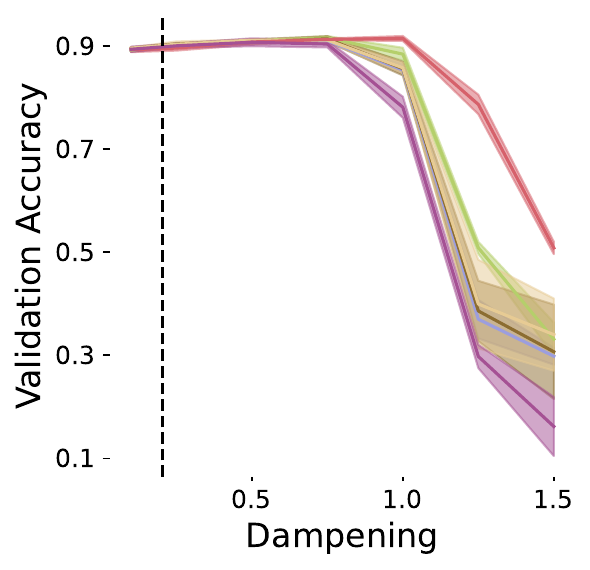}
    \includegraphics[width=.32\textwidth]{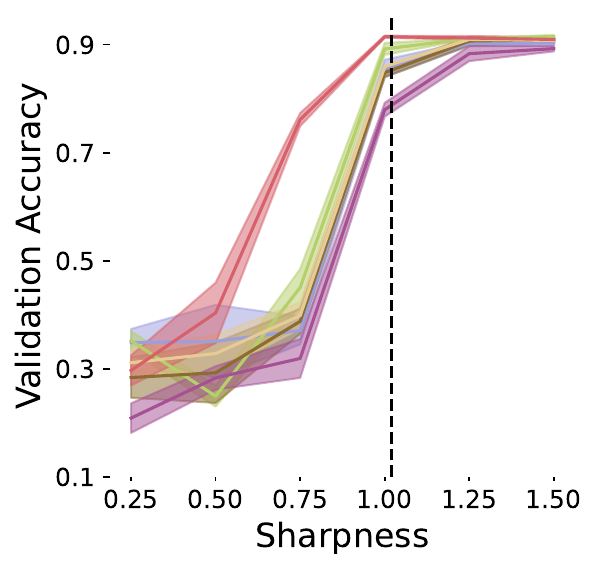}
    \includegraphics[width=.32\textwidth]{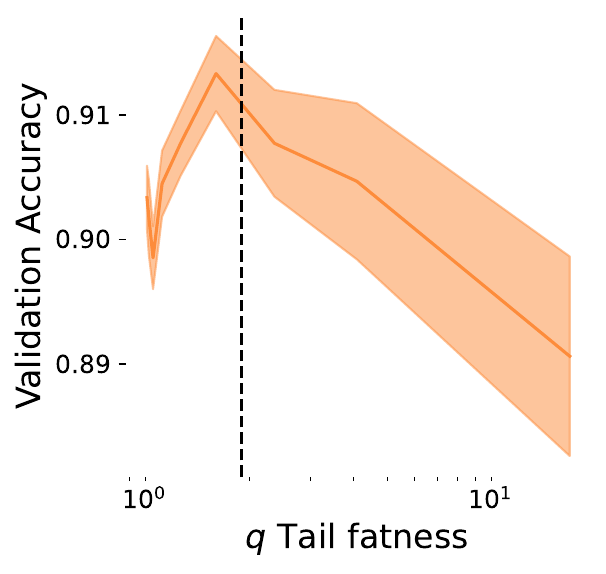}
    
    \includegraphics[width=1.\textwidth]{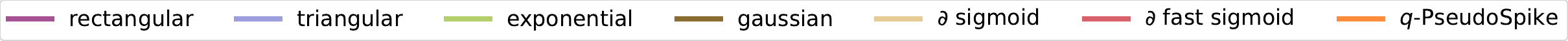}

    \caption{\textbf{Our stability-based theory predicts optimal SG features on the LIF network.} We compare how the features of SG shape predicted by our method stand up against other experimental choices. We conduct the analysis on the LIF network for the sl-MNIST task. Panel (a) shows the performance for different dampening values while setting the sharpness to 1, and vice versa in panel (b). The dashed vertical lines show our theoretical predictions for the exponential SG, (III) for the dampening ($\gamma=0.20\pm 0.02$) and (IV) for the sharpness ($\beta=1.02\pm 0.17$), which agree with the experiments. Dampenings lower than 1 improve performance while the pattern is the opposite for sharpness. Panel (c) shows the performance for different tail-fatness values on the $q$-PseudoSpike SG with $\beta=\gamma=1$. The theoretical prediction gives a close to optimal $q=1.898\pm 0.002$, whereas the best experimental result is $q=1.56$. These findings suggest that our stability-based method predicts good SG features before training, thereby reducing the need for time-consuming hyper-parameter search. }
    \label{fig:sensitivity_to_dsf}
\end{figure*}

\subsection{Our stability-based theory predicts optimal SG features on the LIF network}
\label{sec:heavy_tails}

We compare experimentally the performance of a range of values of dampening, sharpness and tail-fatness and we assess how they compare to our theoretical prediction. Fig.~\ref{fig:sensitivity_to_dsf} shows the accuracy of the LIF network on the sl-MNIST task. Each SG has its tail decay: inverse quadratic for the $\partial$ fast-sigmoid, no tail for the triangular and rectangular, and exponential decays for the rest. Low dampening and high sharpness are preferred by all SG. Interestingly, the accuracy of the $\partial$ fast-sigmoid degrades less with suboptimal $\gamma,\beta$. The vertical dashed lines are predicted by our theoretical method, condition (III) for the dampening and (IV) for the sharpness of an exponential SG. We observe that they find $\gamma,\beta$ with high experimental accuracies. This supports the claim that reducing hyper-parameter search of dampening and sharpness is possible. We use our $q$-PseudoSpike SG to study the dependence with the tail-fatness, panel (c) Fig.~\ref{fig:sensitivity_to_dsf}. All tail-fatness values perform reasonably well, with a maximum at $q=1.56$, smaller than the $q=2$ of the $\partial$ fast-sigmoid. Interestingly our theoretical solution gives a $q=1.898 \pm 0.002$, surprisingly close to the experimental optimum.

\section{Generalizing to more neurons}

It is definitely of interest to be able to generalize this initialization strategy to different architectures and neuron definitions. For that reason, we show in App.~\ref{app:mulreset} the effect that changing the reset definition has on the constraints when the same desired conditions of stability are applied.

Additionally, we compare with the existing results of DECOLLE \cite{decolle} and fluctuations-driven initialization \cite{rossbroich2022fluctuation}. In both cases we used the official implementation\footnote{\url{https://github.com/nmi-lab/decolle-public}}\footnote{\url{https://github.com/fmi-basel/stork}} as referenced in each work, and added the modifications described below. DECOLLE has already what we would consider a stable initialization, given that DECOLLE's LIF has a threshold at zero, and since it uses a feedforward network with an input weight matrix with zero mean, the network tends to have an initial firing rate of 0.5, hitting the most sensitive part of the SG the most often. We compare the default initialization with two scenarios: in the first one we encourage a firing rate of 0.5 in the first epoch of training with a regularization loss, that we remove thereafter; in the second scenario we encourage a sparser firing rate of 0.158 in the first epoch that we remove thereafter. Notice that our firing rate is the mean number of ones in the tensor that contains all the output spikes in a mini batch. As shown in Tab.~\ref{tab:decolle}, encouraging any firing rate with a regularization loss seems to benefit the validation loss, but only encouraging 0.5 seems to benefit the test loss. However the improvements are small, possibly because DECOLLE has a local learning rule, and therefore should avoid the exponential explosion problem by design.

We show in Tab.~\ref{tab:fluctuations} our results compared to the fluctuations-driven initialization (FDI). FDI is based on choosing a parameter $\xi$ between one and three, that determines the variance of the input weights, and results in sparse but not too sparse spiking activity. In deeper networks, FDI outperforms Kaiming initialization \cite{he2015delving}, which is sparser in activity. However, FDI is compared to Kaiming with $\xi=1$ at initialization in the main experiments, the least sparse option within $\xi$ range. Notice that $\xi=1$ corresponds to a firing rate of $\rho=0.158$, which is the reason we chose it in the paragraph above. Moreover, Figure 4 in \cite{rossbroich2022fluctuation} shows a peak in performance for even higher firing rates at initialization, outside the one to three range for the deepest choice of network.  However, since \cite{rossbroich2022fluctuation} fixes the threshold to one and the mean input weights to zero, it can only achieve firing rates of $\rho=0.5$ by increasing the standard deviation of the input weights to values that would render learning unstable. Using \cite{rossbroich2022fluctuation} notation, we propose what we call the mean-driven initialization in the table, to achieve what we consider stable firing rates of $\rho=0.5$ and standard deviation of the gradient of one at initialization. For that, we set the mean voltage as the mean threshold, with an input weight of $\mu_W=\theta/n\nu\overline{\epsilon}$ and $\sigma^2_W=1/n\nu\hat{\epsilon}$ with $\theta=1$ the neuron firing threshold, $n$ the number of presynaptic neurons, and $\nu,\hat{\epsilon}$ are task specific variables that represent expected input activity, that we take as given by the official implementation, except on \mbox{CIFAR-10} deep network, where we set them to one, since it led to a wider performance improvement. Notice that we do a learning rate grid search on the fluctuations-driven initialization, and we take that learning rate for the mean-driven initialization without further grid search. We can see in the table that the mean-driven stable initialization can often match and outperform the fluctuations-driven initialization in deeper LIF architectures. Notice also that the final firing rate is in both cases of 0.01, since an activity regularization loss term was implemented in the official FDI implementation.

\begin{table}[]
\centering
\begin{tabular}{lc}
\textbf{Model} & \textbf{Error} \\
\toprule
DECOLLE test & $7.73 \pm 1.18\%$ \\
DECOLLE test 0.5 & $\textbf{7.55} \pm \textbf{1.22\%}$ \\
DECOLLE test 0.158 & $7.99 \pm 2.30\%$ \\
\\
DECOLLE val & $4.03 \pm 1.07\%$ \\
DECOLLE val 0.5 & $\textbf{3.81} \pm \textbf{1.09\%}$ \\
DECOLLE val 0.158 & $\textbf{3.81} \pm \textbf{1.09\%}$ \\
 \\
DECOLLE \cite{decolle} & $4.46 \pm 0.16\%$ \\
SLAYER \cite{shrestha2018slayer} & $6.36 \pm 0.49\%$ \\
C3D \cite{tran2015learning} & $5.46 \pm 1.06\%$ \\
IBM EEDN \cite{amir2017low} & $8.23\%$ \\
\end{tabular}
\caption{\textbf{DECOLLE Error at the DvsGesture task.} Whenever DECOLLE has no citation is our run of the official implementation with different four seeds. Also notice that in the original implementation the test set was evaluated as a validation set during training, making their reported score misleading. Here we report both test and validation, for the original settings and compare against the scenario where a firing rate of 0.5 is encouraged during one initial epoch through a mean squared error regularization loss, and against a sparser scenario of a target firing rate of 0.158. Both regularizations have a positive effect on validation performance, but only encouraging a high firing rate seems to have an effect that persists on the test set. However the effect is not statistically significant, probably given that DECOLLE is trained with a local loss, and therefore the exponential composition with depth is not an issue. Despite encouraging different firing rates at the beginning of training and none thereafter, all achieve the same firing rate of 0.11 at the end of training.} \label{tab:decolle}
\end{table}

\begin{table}[]
\centering
\begin{tabular}{llcc}
\textbf{Dataset} & \textbf{Model} & \textbf{\shortstack{Test \\ Accuracy \\ \tiny shallow}} & \textbf{\shortstack{Test\\Accuracy \\ \tiny deep}} \\
\toprule
SHD &  &  &  \\
& Mean-driven & $80.0 \pm 3.0$  & $\textbf{83.4} \pm \textbf{3.0}$ \\
& Fluct.-driven & $83.0 \pm 2.6$ & $82.5 \pm 2.3$ \\
& Fluct.-driven \cite{rossbroich2022fluctuation} & $82.7 \pm 1.1$ & $80.9 \pm 1.2$ \\
& Kaiming \cite{rossbroich2022fluctuation, he2015delving} & $\textbf{83.1} \pm \textbf{1.2}$ & $4.5 \pm 0.0$ \\
CIFAR-10 &  &  &  \\
& Mean-driven & $\textbf{63.5} \pm \textbf{0.3}$  & $\textbf{69.2} \pm \textbf{1.1}$ \\
& Fluct.-driven & $63.4 \pm 0.6$  & $64.5 \pm 0.5$ \\
& Fluct.-driven \cite{rossbroich2022fluctuation} & $62.4 \pm 0.3$  & $65.6 \pm 1.3$ \\
& Kaiming \cite{rossbroich2022fluctuation, he2015delving} & $59.5 \pm 0.8$ & $10.0 \pm 0.0$ \\
DVS &  &  &  \\
& Mean-driven & $85.0 \pm 2.0$ & $88.0 \pm 0.6$ \\
& Fluct.-driven & $\textbf{88.3} \pm \textbf{1.1}$ & $\textbf{89.3} \pm \textbf{1.7}$ \\
& Fluct.-driven \cite{rossbroich2022fluctuation} & $86.7 \pm 1.2$ & $86.4 \pm 1.7$ \\
& Kaiming \cite{rossbroich2022fluctuation, he2015delving} & $54.6 \pm 37.1$ & $9.1 \pm 0.0$ \\
\end{tabular}
\caption{\textbf{Fluctuations-driven against mean-driven initializations of LIF neurons.} We implement our stable initialization in the spiking implementation from \cite{rossbroich2022fluctuation}. We report \cite{rossbroich2022fluctuation} results and our run for different seeds and smaller batch sizes to fit our resources. Kaiming results in the sparsest initialization, that leads to the worst results, especially for deeper networks. The fluctuations-driven initialization results in a firing rate of 0.158 at initialization while our mean-driven, results in a firing rate of 0.5 at initialization and a standard deviation of one of the gradient. We see that deeper networks can benefit of high firing rates at initialization, while achieving the same firing rate by the end of training.}\label{tab:fluctuations}
\end{table}

\section{Discussion and Conclusions}

Our method based on stabilizing forward and backward pass, resulted in improved accuracy over the baseline and it was able to predict optimal dampening, sharpness and tail-fatness before training. 
Our findings are coherent with the line of research that has established that stabilizing gradients and representations at initialization results in better performance \cite{glorot2010understanding, orthogonal_initialization, he2015delving, roberts2022principles, defazio2022scaling, bengio1994learning, hochreiter1997long, hochreiter2001gradient, arjovsky2016unitary, pascanu2013difficulty}. Moreover it gives an initial reply to the question raised by
\cite{surrogate2019, zenke2021remarkable}, which asked  for a theoretical justification of initialization and SG choice for Spiking Neural Networks. With a similar intention, \cite{rossbroich2022fluctuation} proposed an approach that guarantees sparsity of activity at initialization to pick the weights distribution at initialization, resulting in improved accuracy. Our method differs from theirs in that it starts from a principle of stability to derive constraints, instead of a principle of sparsity. It differs also in that we use it to define the SG shape at initialization, not only the weights distribution, and we show mathematically how weights initialization is intertwined to the SG shape choice. Our results suggest that a tedious hyper-parameter grid-search can be often avoided by making use of sound and established principles of learning stability.

One of the stability conditions was designed to hit the most sensitive part of an SG, its center, which resulted in a high frequency requirement at initialization. This is very uncommon in the Neuromorphic literature, since sparsity brings large energy gains \cite{henderson2020towards,blouw2019benchmarking, 9395703,taulsnn, rossbroich2022fluctuation}.
However, the energy gains of SNNs also come from their binary activity. A matrix-vector multiplication, with a $\mathbb{R}^{m\times n}$ matrix, has an energy cost of $mnE_{MAC}$ for a real vector, and of $mn\rho E_{AC}$ for a binary vector, where $\rho$ is the Bernouilli probability of the binary vector, and in our case the neuron firing rate, and $E_{AC}, E_{MAC}$ are the energies of an accumulate and a multiply-accumulate operation \cite{yin2021accurate, hunger2005floating}. Since MAC are more costly than AC, 31 times on a $45$nm complementary metal–oxide–semiconductor \cite{yin2021accurate, horowitz20141}, we have energy savings with any $\rho$, e.g. when all neurons fire ($\rho=1$) and when they fire half of the time steps ($\rho=1/2$). This gain does not depend on the simulation speed, since it compares a spiking and an analogue computation, at the same computation speed.
Typically requiring more sparsity through a sparsity encouraging loss term, leads to a measurable decrease in performance \cite{zenke2021remarkable, rossbroich2022fluctuation}. However we observed that it is actually possible to achieve higher performance with higher sparsity, by starting with a strong firing rate at initialization, since their synergy acts as a regularization mechanism. This was possible also because the sparsity encouraging loss term was introduced gradually, and because its contribution was kept comparable to the task loss towards the end of training. We observed a similar behavior when we set DECOLLE \cite{decolle} and fluctuations-driven \cite{rossbroich2022fluctuation} architectures to have a high firing rate at initialization: better final performance was obtained by deeper networks with the same final sparsity. This also shows that stability arguments might be of relevance for a wide variety of spiking architectures.

Finally, we observed that the more complex the task is and the more complex the network to train is, the more drastic is the difference in performance of different SG shapes. It is known that learning is possible with a wide variety of SG shapes \cite{zenke2021remarkable} and the community has not yet settled for one shape or one method to reliably choose which SG to use in each case \cite{surrogate2019}. We showed how to apply a well known stability principle to the forward and backward pass of the simplest Spiking Neural Network, the LIF, as a starting point, but we think that the principles of good Neuromorphic initialization can be further elaborated, in order to tackle more complex tasks and networks.

\section{Acknowledgements}

Luca Herranz-Celotti was supported by the Natural Sciences and Engineering Research Council of Canada through the Discovery Grant from professor Jean Rouat, and by CHIST-ERA IGLU. We thank Compute/Digital Research Alliance of Canada for the clusters used to perform the experiments and NVIDIA for the donation of two GPUs. We thank Wolfgang Maass for the opportunity to visit the Institute of Theoretical Computer Science, Guillaume Bellec, Darjan Salaj and Franz Scherr, for their invaluable insights on learning with surrogate gradients, and Maryam Hosseini, Ahmad El Ferdaoussi and Guillaume Bellec for their feedback on the article.

\bibliographystyle{unsrt} 
\bibliography{references}



\clearpage
\renewcommand{\thesection}{\Alph{section}}
\renewcommand{\theHsection}{A\arabic{section}}

\beginsupplement



\section*{Appendix}

\section{More Training Details}
\label{app:trainingdetails}

We collect the training hyper-parameters in the following table.

\begin{table}[h]
\centering
\begin{tabular}{rlll}
& sl-MNIST & SHD &  PTB \\ \hline
\\
batch size & 256 & 256 & 32 \\
weight decay &0.0 & 0.1 &  0.1 \\
gradient norm & 1.0 & 1.0 & 1.0 \\
\shortstack{train/val/test \\ \ } & \shortstack{45k/5k/10k\\ samples } & \shortstack{8k/1k/2k \\ samples }  &   \shortstack{930k/74k/82k \\ words} \\
learning rate & $3.16 \cdot 10^{-4}$ &  $3.16 \cdot 10^{-4}$ &  $3.16 \cdot 10^{-5}$ \\
layers width & 128, 128 & 256, 256 &  1700, 300 \\
label smoothing & 0.1 & 0.1 & 0.1 \\
time step repeat & 2 & 2 & 2 \\
SELT factor & 0.8 & 0.436 & 4.595 \\
\end{tabular}
\end{table}

The learning rates were chosen after a grid search fixing dampening and sharpness to 1.  The learning rates considered are in the set $\{10^{-2},3.16 \cdot10^{-3}, 10^{-3}, 3.16 \cdot10^{-4}, 10^{-4}, 3.16 \cdot10^{-5},10^{-5 }\}$. The results of the grid search are reported in figure \ref{fig:task_net_dependence}. The learning rate chosen for the rest of the paper was the one that made all the shapes perform reasonably well, rectangular included. This mostly resulted in a suboptimal learning rate only for the derivative of the fast sigmoid, which still out-performed the rest in the sl-MNIST and SHD, and performed comparatively on the PTB. 

We train with crossentropy loss, the AdaBelief optimizer \cite{zhuang2020adabelief}, Stochastic Weight Averaging \cite{swa2018} and Decoupled Weight Decay \cite{loshchilov2018decoupled}. 
For the PTB task, the input passes through an embedding layer before passing to the first layer, and the output of the last layer is multiplied by the embedding to produce the output, removing the need for the readout
\cite{wozniak2020deep, radford2018improving}.
Notice that we do not implement forced refractory periods that would prevent the neuron from firing too fast, as sometimes done in the neuromorphic literature, since we want to reduce the non differentiable steps in the system. Thus, $\rho=1$ is possible if the inputs are strong and frequent enough.

\section{Neuron Model Complexity}

The energy consumed per layer can be used as a metric of neuron complexity, as done in \cite{yin2021accurate, hunger2005floating}.

\begin{table}[h]
\begin{center}
\begin{tabular}{ll}
\textbf{\shortstack{Neural \\ model}} & \textbf{\shortstack{Energy (Complexity) \\ \hspace{1cm}}}\\
\hline\\
\textbf{LIF} & \shortstack{$(mnp_{l-1}+nnp_{l})E_{AC}+nE_{MAC}$ }\\
\textbf{ALIF} & \shortstack{$(mnp_{l-1}+nnp_{l}+2np_{l})E_{AC}+3nE_{MAC}$ } \\
\textbf{LSTM} & \shortstack{$4(mn+nn)E_{MAC}+17nE_{MAC}$ } \\
\textbf{sLSTM} & \shortstack{$4(mnp_{l-1}+nnp_{l})E_{AC}+3np_{l}E_{AC}$ } 
\end{tabular}
\end{center}
\caption{\textbf{Neuron complexity.} We use the energy consumed per layer as a metric of neuron complexity \cite{yin2021accurate, hunger2005floating}. We use  $n=n_l$ and $m=n_{l-1}$ as the width of the layer and its input, $p_{l}$ for the  firing rate of the layer $l$. $E_{MAC}$ is the energy cost of a multiply-accumulate operation and $E_{AC}$ of an accumulate operation. As shown, $ALIF$ always results in a larger number of operations and energy consumption than $LIF$. For large networks, $n,m\gg1$, the square terms dominates, and the $sLSTM$ results in 4 times more energy consumption. 
}
\label{tab:complexities}
\end{table}

\newpage

\section{List of Surrogate Gradients shapes}
\label{app:surrogate}

We list here the shapes that we used in this article as surrogate gradients.

\begin{table}[h]
\centering
\small
\begin{tabular}{llll}
 & SG name  & $f(v)$  \\ \\ 
\textcolor{ctr}{\rule{1cm}{2pt}} & \textbf{triangular} & $ \max(1-|v|,0)$  \\ 
\textcolor{cex}{\rule{1cm}{2pt}} & \textbf{exponential} & $e^{-2|v|}$ \\ 
\textcolor{cga}{\rule{1cm}{2pt}} & \textbf{gaussian}  & $ e^{-\pi v^2}$ \\ 
\textcolor{csi}{\rule{1cm}{2pt}} & \textbf{$\partial$ sigmoid} & $4\ sigmoid(4\ v)\left(1-sigmoid(4\ v)\right)$ \\ 
\textcolor{cfa}{\rule{1cm}{2pt}} & \textbf{$\partial$ fast-sigmoid} & $\frac{1}{(1+|2v|)^2}$ \\ 
\textcolor{cre}{\rule{1cm}{2pt}} & \textbf{rectangular} &  $\mathbb{1}_{|v|<\frac{1}{2}}$ \\ \\ 
\textcolor{cnt}{\rule{1cm}{2pt}} &
\shortstack{\textbf{$q$-PseudoSpike} \\ ($q>1$)} & $\frac{1}{(1+\frac{2}{q-1}|v|)^q}$ \\ \\
\end{tabular}
\caption{\textbf{Mathematical definitions of the surrogate gradients studied in this article.}
Our Heaviside activation $\sigma(v)=\tilde{H}(v)$, where $v$ is the centered voltage, has the SG $\sigma'(v) = \gamma f(\beta\cdot v)$, where $\beta$ is the SG sharpness, $\gamma$ the SG dampening, and $f$ is the shape of choice. The constants, are chosen for the SG to have a maximal value of $1$ and an area under the curve of $1$.}
\end{table}


\section{Detailed derivation of the conditions}

\label{sec:conditionsdetails}

We derive the constraints on the hyper-parameters that will lead the LIF to meet the conditions proposed at initialization. The LIF we will be using
is defined by

\begin{align}
    \boldsymbol{y}_t = \boldsymbol{\alpha}_{decay} \boldsymbol{y}_{t-1}(1-\boldsymbol{x}_{t-1}) + \boldsymbol{i}_{t}
\end{align}

\noindent where $\boldsymbol{i}_{t}=W_{rec}\boldsymbol{x}_{t-1} + W_{in}\boldsymbol{z}_t + \boldsymbol{b}$, as described in the main text, and the multiplicative factor $(1-\boldsymbol{x}_{t-1})$ represents the reset mechanism.

\subsection{Recurrent matrix mean sets the firing rate (I)}

\label{app:means}

We show how condition (I) leads to a constraint on the mean of the recurrent connectivity with a LIF neuron model, that will lead the network to meet that condition at initialization.

\begin{lemma}\label{thm:cI}
Applying condition (I), which states that we want
$Median[v] = 0$, to an LIF network, and further assuming
$\overline{w}_{in} = 0, b = 0$, the approximation $Mean[v] \approx Median[v]$,
and constant $\overline{i}_t$ over time, it results in the constraint

\begin{align}
    \overline{w}_{rec}=\frac{1}{n_{rec}-1}(2-\overline{\alpha}_{decay})\overline{\vartheta}
\end{align}

\end{lemma}

\begin{proof}

First we show that $Median[v]=0\implies Mean[x]=1/2$, where $x = \tilde{H}(v)$. In equation \ref{eq:marg} we write the marginal distribution of $p(x)=\int p(x|v)p(v)dv$, and the double integral is represented with one integration symbol. Then, we notice that $x$ has a deterministic dependence on $v$, $x=H(v)$, which proprbabilistically is described by the delta function $p(x|v)=\delta(x - H(v))$. Then, we integrate over $x$, and in the last equation we notice that integrating with respect to the Heaviside is equivalent to restricting the integration limits from zero to infinity.

\begin{align}
    Mean[x] =& \int xp(x) dx \\
    =& \int xp(x|v)p(v)dxdv  \label{eq:marg}\\
    =& \int x p(v)\delta(x - H(v))dxdv\\
    =& \int p(v)dv H(v)\\
    =& \int_0^{\infty} p(v)dv
\end{align}

If $Median[v]=0$, half of it's probability mass is on each side of $0$, so the last integral is equal to $1/2$, QED.

Since working with medians is mathematically harder than working with means, we assume that $Mean[v]\approx Median[v]$, with the caveat that it will make the result approximate. To justify that they are similar, it can be shown that for a unimodal distribution $v\sim p(v)$ with the first two moments defined, we have $|Mean[v]-Median[v]|\leq\sqrt{0.6Var[v]}$ \cite{basu1997mean}. We will proceed with this approximation in mind, and we will continue the development with means and not with medians.

We use the notation $\overline{x}=Mean[x]$ interchangeably.
We calculate how the mean of the voltage elements is propagated through time, assuming the mean input current to remain constant over time $\overline{i}_t = \overline{i}$ at initialization, to simplify the mathematical development, and assuming per condition (I), that $\overline{x} = \overline{1-x}=1/2$ we have

\begin{align}
    \overline{y}_t 
    =&\overline{\alpha}_{decay} \overline{(1-\boldsymbol{x}_{t-1})}\overline{y}_{t-1} + \overline{i} \\
    =&\frac{1}{2}\overline{\alpha}_{decay} \overline{y}_{t-1} + \overline{i} \\
    =&\frac{1}{2}\overline{\alpha}_{decay} \Big(\frac{1}{2}\overline{\alpha}_{decay} \overline{y}_{t-2} + \overline{i}\Big) + \overline{i} \\
    =&\frac{1}{2^{t-1}}\overline{\alpha}_{decay}^{t-1}\overline{y}_{1} + \Big(\sum_{t'=0}^{t-2}\frac{1}{2^{t'}}\overline{\alpha}_{decay}^{t'}\Big)\overline{i} \\
    =&\frac{1}{2^{t-1}}\overline{\alpha}_{decay}^{t-1}\overline{y}_{1}  + \frac{1-\frac{1}{2^{t-1}}\overline{\alpha}_{decay}^{t-1} }{1-\frac{1}{2}\overline{\alpha}_{decay}}\overline{i} 
\end{align}

\noindent where we used the fact that the same LIF definition applies to different time steps, the geometric series formula, and the fact that for independent random variables $E[XY] =E[X]E[Y]$. For $t\rightarrow\infty$ and using  $0<\overline{\alpha}_{decay}<1$

\begin{align}
    \overline{y}_t =&
    \frac{1 }{1-\frac{1}{2}\overline{\alpha}_{decay}}\overline{i} \\
    \overline{y}_t-\overline{\vartheta} =&
    \frac{1 }{1-\frac{1}{2}\overline{\alpha}_{decay}}\overline{i} - \overline{\vartheta} 
\end{align}

Assuming we want this condition to hold independently of the dataset, we set $Mean[W_{in}]=0$, and assuming that we do not want to promote this behavior with fixed internal currents, but with the recurrent activity instead, then $\boldsymbol{b}=0$. 

We remark that we denote $Mean[W\boldsymbol{x}]$ as the mean vector whose element $i$ is

\begin{align}\label{eq:indepx}
    Mean[W\boldsymbol{x}]_i=& Mean[\sum_{\substack{j=1\\j\neq i}} w_{ij}x_j] \\
    =& \sum_{\substack{j=1\\j\neq i}} Mean[ w_{ij}x_j]   \\
    =& \sum_{\substack{j=1\\j\neq i}} Mean[ wx]  \\
    =& (n_{rec}-1) Mean[ wx]   
\end{align}

\noindent where the condition $j\neq i$ in the summand reminds that neurons are not connected to themselves in our recurrent architecture. In the first equality, the index $i$ denoting the element in the vector, is equivalent as choosing the row $i$ of $W$, so it is not necessary to specify it outside the square brakets. The equality before the last one is a consequence of considering any neuron as mutually independent to any other at initialization, as done by \cite{glorot2010understanding, he2015delving}, and that justifies dropping the indices. Since $w$ and $x$ are statistically independent random variables, $Mean[ wx]=Mean[ w]Mean[ x]$.

Then, 

\begin{align}
    Mean[y_t-\vartheta] =&
    \frac{1 }{1-\frac{1}{2}\overline{\alpha}_{decay}} \Big((n_{rec}-1)\overline{w}_{rec}\overline{x}_{t-1}\Big) - \overline{\vartheta}\\
    0=&
    \frac{1 }{1-\frac{1}{2}\overline{\alpha}_{decay}} (n_{rec}-1)\overline{w}_{rec}\overline{x}_{t-1} - \overline{\vartheta}\\
    0=&
    \frac{1}{1-\frac{1}{2}\overline{\alpha}_{decay}} (n_{rec}-1)\overline{w}_{rec}\frac{1}{2} - \overline{\vartheta}\\
    (n_{rec}-1)\overline{w}_{rec}=&
    (2-\overline{\alpha}_{decay})\ \overline{\vartheta}\\
    \overline{w}_{rec}=&
    \frac{1}{n_{rec}-1}\Big(2-\overline{\alpha}_{decay}\Big)\overline{\vartheta} \label{appeq:CI}
\end{align}

\noindent where in the second line we applied condition (I) in the form of $Mean[v_t]\approx Median[v_t]=0$, so $Mean[y_t-\vartheta]=0$, and in the third line we applied again condition (I), $\overline{x}_t=1/2$.
In the main text we turn $\overline{\vartheta}, \overline{\alpha}_{decay}\rightarrow \vartheta, \alpha_{decay}$, since here we consider the more general case where those are as well random variables, and we simplify it in the main text for cleanliness, assuming they are constant.

\end{proof}

We therefore found a constraint on the mean of the recurrent matrix initialization, that leads the LIF network to satisfy condition I at initialization. The constraint is equation \ref{appeq:CI} with $\overline{w}_{in}=0$, and  $\boldsymbol{b}=0$.

\subsection{Recurrent matrix variance can  make recurrent and input voltages comparable (II)}
\label{app:ineqrec}

We apply condition (II) to the LIF network, that gives us a constraint that the recurrent matrix has to meet at initialization for the condition to be true.

\begin{lemma}\label{thm:cII}
Applying condition (II), which states that we want $Var[W_{rec}x_{t-1}] =  \ Var[W_{in}z_t]$, to an LIF network, and further assuming $\overline{x}=1/2$, and $\overline{w}_{in}=0$, it results in the constraint

\begin{align}\label{eq:condition_II}
    &Var[w_{rec}]  =  2(Var[z_t] + \overline{z}_t^2)\frac{n_{in}}{n_{rec}-1}Var[w_{in}] - \frac{1}{2}\overline{w}_{rec}^2
\end{align}
\end{lemma}

\begin{proof}
The second condition, is that the recurrent and the input contribution to the variance need to match

\begin{align} 
    Var[W_{rec}x_{t-1}] =&  \ Var[W_{in}z_t]
\end{align}

\noindent where the variance is computed at each element, after the matrix multiplication is performed, following the method described in \cite{glorot2010understanding, he2015delving}. Similarly to what we did for the means in equation \ref{eq:indepx}, the matrix multiplication contributes to the scalar variance of neuron $i$ as

\begin{align}\label{eq:indepxvar}
    Var[W\boldsymbol{x}]_i=&Var[\sum_{j=1} w_{ij}x_j] \\
    =& \sum_{j=1} Var[ w_{ij}x_j] \\
    =& \sum_{j=1} Var[ wx]  \\
    =& n_W Var[ wx]
\end{align}

The second and third equality are a consequence of considering any neuron as mutually independent to any other at initialization, as done by \cite{glorot2010understanding, he2015delving}, and that justifies that the variance of the sum is the sum of the variances, and it justifies dropping the indices, to mean that the statistics are the same for each element. The number $n_W$ stands for the number of inputs that a neuron $i$ has through $W$, in the case of $W_{in}$, $n_W=n_{in}$, while in the case of $W_{rec}$, we have $n_W=n_{rec}-1$, since in our recurrent network, neurons are not connected to themselves.

Therefore the vector-wise condition II is equivalent to the element-wise

\begin{align}
    (n_{rec}-1)Var[w_{rec}x_{t-1}] =& \  n_{in}Var[w_{in}z_t]
\end{align}

Since the time dimension is averaged out, the time axis can be randomly shuffled, and the LIF activity is indistinguishable from a Bernouilli process through the mean and variance of the activity. Therefore if $\overline{x}_t=p$, we have $Var[x_t]=p(1-p)$ when averaged over time, with $p$ the probability of firing. Therefore it is as well true that $\overline{x_t^2}=Var[x_t] + \overline{x}_t^2 =p$.

We apply the fact that for independent $w, x$

\begin{align}
    Var[wx] = \overline{w^2}\ \overline{x^2} - \overline{w}^2\overline{x}^2
\end{align}

\noindent and assuming $\overline{w}_{in}=0$ and $p=1/2$ we have

\begin{align}
    Var[w_{rec}x_{t-1}] =& (Var[w_{rec}] + \overline{w}_{rec}^2)p-\overline{w}_{rec}^2p^2 \\ 
    =& \frac{1}{4}(2Var[w_{rec}] + \overline{w}_{rec}^2) \\
    Var[w_{in}z_t] =& (Var[z_t] + \overline{z}_t^2)Var[w_{in}]
\end{align}

Substituting in equation \ref{eq:condition_II} implies 

\begin{align}
    &\frac{1}{4}(2Var[w_{rec}] + \overline{w}_{rec}^2) =  (Var[z_t] + \overline{z}_t^2)\frac{n_{in}}{n_{rec}-1}Var[w_{in}] \\
    &Var[w_{rec}]  =  2(Var[z_t] + \overline{z}_t^2)\frac{n_{in}}{n_{rec}-1}Var[w_{in}] - \frac{1}{2}\overline{w}_{rec}^2 \label{appeq:CII}
\end{align}

\end{proof}

Therefore condition (II) led us to the constraint that $W_{rec}$ has to meet at initialization, equation \ref{appeq:CII}, for the condition to be true. The final equation further assumes that  $\overline{w}_{in}=0$ and $p=1/2$.

\subsection{SG dampening controls gradient maximum (III)}

\label{app:sgmax}

We apply condition (III) to the LIF network, which gives us a constraint that the dampening has to meet at initialization for the condition to be true.

\begin{lemma}\label{thm:cIII}
Applying condition (III), which states that we want $Max[\frac{\partial}{\partial \theta}y_t] = Max[\frac{\partial}{\partial \theta}y_{t-1}]$, to an LIF network, and assuming that (1) $\sigma'$ and $\frac{\partial}{\partial \theta}y_{t-1}$ are statistically independent and (2) we do not pass the gradient through the reset, it results in the constraint  

\begin{align}
    \gamma=&  
    \frac{1}{(n_{rec}-1)\hat{w}_{rec}}\Big( 1 -\hat{\alpha}_{decay} -\xi \cdot n_{in} \hat{w}_{in} \gamma_{in} \Big)
\end{align}

\noindent where $\xi$ is zero for the first layer and it's one for the other layers in the stack.

\end{lemma}

\begin{proof}

We want the maximal value of the gradient to remain stable, without exploding, when transmitted through time and through different layers

\begin{align}
    Max[\frac{\partial}{\partial \theta}y_t] = Max[\frac{\partial}{\partial \theta}y_{t-1}]
\end{align}

\noindent where when we write $\partial/\partial \theta$, we use $\theta$ as a placeholder for any quantity that we want to propagate through gradient descent.
Taking the derivative of the LIF definition and stopping the gradient from going through the reset we have 

\begin{align}\label{eq:difLIF}
    \frac{\partial}{\partial \theta}y_t =& \alpha_{decay} \frac{\partial}{\partial \theta}y_{t-1}(1-x_{t-1}) + W_{rec}\frac{\partial}{\partial \theta}x_{t-1} + \xi W_{in}\frac{\partial}{\partial \theta}z_{t} 
\end{align}

Here we introduce the symbol $\xi\in \{0,1\}$, where $\xi=1$ is used when $z_t$ comes from a trainable layer below, and $\xi=0$ when $z_t$ represents the data. We consider as well that

\begin{align}
    \frac{\partial}{\partial \theta}z_{t} =& \frac{\partial}{\partial \theta}\tilde{H}_{in}(y_{t}^{in}-\vartheta_{in})= \sigma'_{in}\frac{\partial}{\partial \theta}y_{t}^{in} \\
    \frac{\partial}{\partial \theta}x_{t-1} =& \frac{\partial}{\partial \theta}\tilde{H}(y_{t-1}-\vartheta)=\sigma'\frac{\partial}{\partial \theta}y_{t-1}
\end{align}

\noindent where  $\tilde{H}_{in},y_{t}^{in},\vartheta_{in}$ are the Heaviside, the voltage and the threshold of the layer below, $\sigma'=\frac{\partial\tilde{H}}{\partial v}$ is the surrogate gradient, and $\sigma'_{in}$ is the surrogate gradient from the layer below. Substituting in equation \ref{eq:difLIF}, then

\begin{align}
    \frac{\partial}{\partial \theta}y_t =& \alpha_{decay} \frac{\partial}{\partial \theta}y_{t-1}(1-x_{t-1}) + W_{rec}\sigma'\frac{\partial}{\partial \theta}y_{t-1} \\
    &+ \xi W_{in}\sigma'_{in}\frac{\partial}{\partial \theta}y_{t}^{in}  \label{appeq:derivativelif}
\end{align}

We use  $Max$ and $Min$ in a statistical ensemble sense, as the maximum/minimum value that a variable could take if sampled over and over again

\begin{align}
    Max[X] =& \sup_{x\sim p(x)} x \\
    Min[X] =& \inf_{x\sim p(x)} x 
\end{align}

With this definition, if $X,Y$ are independent random variables  $Max[X+Y]=Max[X] + Max[Y]$ and if they are positive $Max[XY]=Max[X]Max[Y]$.
We observe, as we did before for the variance and the mean of $Wx$, that

\begin{align}
    Max[Wx] =& n_WMax[wx] \\
    Min[Wx] =& n_WMin[wx]
\end{align}

We take the maximal value of $\frac{\partial}{\partial \theta}y_t$, we make the  assumption that $\sigma'$ and $\frac{\partial}{\partial \theta}y_{t-1}$ are statistically independent,  we use the fact that the highest value that the surrogate gradient can take is given by the dampening factor $Max[\sigma']=\gamma$, we denote as $\gamma_{in}$ the dampening factor of the layer below in the stack, and we take $Max[1-x_{t-1}]=1$:

\begin{align}
    Max[\frac{\partial}{\partial \theta}y_t] =& Max[\alpha_{decay} \frac{\partial}{\partial \theta}y_{t-1}] 
    \nonumber\\&+ (n_{rec}-1)Max[w_{rec}\sigma'\frac{\partial}{\partial \theta}y_{t-1}] \nonumber\\ &+ \xi n_{in}Max[ w_{in}\sigma'_{in}\frac{\partial}{\partial \theta}y_{t}^{in}] \\
    =& Max[\alpha_{decay} ] Max[\frac{\partial}{\partial \theta}y_{t-1}] 
    \nonumber\\&+ (n_{rec}-1)Max[w_{rec}] Max[\sigma'] Max[\frac{\partial}{\partial \theta}y_{t-1}] \nonumber \\
    &+ \xi n_{in} Max[w_{in}] Max[\sigma'_{in}]Max[\frac{\partial}{\partial \theta}y_{t}^{in}]  \\
    =& Max[\alpha_{decay} ] Max[\frac{\partial}{\partial \theta}y_{t-1}]
    \nonumber\\&+ (n_{rec}-1)Max[w_{rec}] \gamma Max[\frac{\partial}{\partial \theta}y_{t-1}] \nonumber \\
    &+\xi n_{in} Max[w_{in}] \gamma_{in}Max[\frac{\partial}{\partial \theta}y_{t}^{in}] 
\end{align}

\noindent where we used the fact that $\sigma'$ is positive in the second equality. We apply condition (III), which states that all maximal gradients are equivalent, and for cleanliness we use the notation $Max[x] = \hat{x}$

\begin{align}
    1 =& \hat{\alpha}_{decay}  +(n_{rec}-1) \hat{w}_{rec} \gamma + \xi n_{in}\hat{w}_{in} \gamma_{in}\\
    \gamma=&  
    \frac{1}{(n_{rec}-1)\hat{w}_{rec}}\Big( 1 -\hat{\alpha}_{decay} -\xi \cdot n_{in} \hat{w}_{in} \gamma_{in} \Big) \label{appeq:CIII}
\end{align}

\noindent where we only had to rearrange terms. 

\end{proof}

We set $\xi=0$ in the main text for readability and because we observed better performance with it.
This final equation \ref{appeq:CIII} gives the value that the dampening has to take to keep the maximal gradient value stable, namely, condition (III) true at initialization.

\subsection{SG sharpness controls gradient variance (IV)}

\label{app:backward}

We apply condition (IV) to the LIF network to constrain the choice of surrogate gradient variance.

\begin{lemma}\label{thm:cIV}
Applying condition (IV), which states that we want $Var[\frac{\partial}{\partial \theta}y_t] = Var[\frac{\partial}{\partial \theta}y_{t-1}]$, to an LIF network, and assuming that (1) we do not pass the gradient through the reset, and (2) zero mean gradients at initialization, it results in the constraint

\begin{align} \label{appeq:CIV}
    \overline{\sigma^{\prime 2}} =&  \frac{1-\frac{1}{2}\overline{\alpha^2}_{decay}-\xi \cdot n_{in}\overline{w_{in}^2}\ \overline{\sigma^{\prime 2}_{in}}}{(n_{rec}-1)\overline{w^2}_{rec} }  
\end{align}

\noindent where $\xi$ is zero for the first layer and is one for the other layers in the stack.

\end{lemma}

\begin{proof}

Condition (IV) states that we want the variance of the gradient to remain stable across time and layers.
Taking the derivative of the LIF we arrive at equation \ref{appeq:derivativelif}:

\begin{align}
    \frac{\partial}{\partial \theta}y_t =& \alpha_{decay} \frac{\partial}{\partial \theta}y_{t-1}(1-x_{t-1}) \nonumber \\&+ W_{rec}\sigma'\frac{\partial}{\partial \theta}y_{t-1} + \xi W_{in}\sigma'_{in}\frac{\partial}{\partial \theta}y_{t}^{in}
\end{align}

Taking the variance and assuming that the monomials in the polynomial are statistically independent, we can consider the variance of the sum to be the sum of the variances:

\begin{align}
    Var[\frac{\partial}{\partial \theta}y_t] =& Var[\alpha_{decay} \frac{\partial}{\partial \theta}y_{t-1}(1-x_{t-1})] \nonumber \\&+ Var[W_{rec}\sigma'\frac{\partial}{\partial \theta}y_{t-1}] \nonumber \\&+ Var[\xi W_{in}\sigma'_{in}\frac{\partial}{\partial \theta}y_{t}^{in}] \\
    Var[\frac{\partial}{\partial \theta}y_t] =& Var[\alpha_{decay} \frac{\partial}{\partial \theta}y_{t-1}(1-x_{t-1})] \nonumber \\&+ (n_{rec}-1)Var[w_{rec}\sigma'\frac{\partial}{\partial \theta}y_{t-1}] \nonumber \\&+ n_{in}Var[\xi w_{in}\sigma'_{in}\frac{\partial}{\partial \theta}y_{t}^{in}]
\end{align}

\noindent where $\xi=0$ if $w_{in}$ connects to the data and $\xi=1$ if it connects to the layer below in the stack. We denote by $\sigma'_{in}$  the surrogate gradient of the layer below.

Assuming gradients $g$ with mean zero, and weights and gradients $w,g$ to be independent random variables at initialization:

\begin{align}
    Var[wg] =& (Var[g] + E[g]^2)(Var[w] + E[w]^2)-E[g]^2E[w]^2 \\
    =& Var[g](Var[w] + E[w]^2) \\
    =&  Var[g] E[w^2]
\end{align}

\noindent which gives

\begin{align}
    Var[\frac{\partial}{\partial \theta}y_t] =& E[\alpha_{decay}^2(1-x_{t-1})^2] Var[\frac{\partial}{\partial \theta}y_{t-1}] \nonumber \\&+ (n_{rec}-1)E[(w_{rec}\sigma')^2]Var[\frac{\partial}{\partial \theta}y_{t-1}] \nonumber \\&+ \xi \cdot n_{in}E[(w_{in}\sigma'_{in})^2]Var[\frac{\partial}{\partial \theta}y_{t}^{in}]
\end{align}

We apply condition IV, we want gradients to have the same variance, irrespective of the time step, or the neuron in the stack, which results in

\begin{align}
    1=& \frac{1}{2}E[\alpha_{decay}^2]  + (n_{rec}-1)E[(w_{rec}\sigma')^2] \nonumber \\&+\xi \cdot n_{in}E[(w_{in}\sigma'_{in})^2]    \\
    1=& \frac{1}{2}E[\alpha_{decay}^2]  + (n_{rec}-1)E[w_{rec}^2]E[\sigma^{\prime 2}] \nonumber \\&+\xi \cdot n_{in}E[w_{in}^2]E[\sigma^{\prime 2}_{in}]  
\end{align}

\noindent where  we used the fact that for independent variables $X, Y$ we have $E[X^pY^q] = E[X^p]E[Y^q]$ in the third and fourth line. Using the notation $E[x]=\overline{x}$, the implied condition on the SG is

\begin{align} 
    \overline{\sigma^{\prime 2}} =&  \frac{1-\frac{1}{2}\overline{\alpha^2}_{decay}-\xi \cdot n_{in}\overline{w_{in}^2}\ \overline{\sigma^{\prime 2}_{in}}}{(n_{rec}-1)\overline{w^2}_{rec} }  
\end{align}

\end{proof}

We therefore found the constraint that the second non-centered moment of the SG has to satisfy, equation \ref{appeq:CIV}, if we want condition IV to hold. We set $\xi=0$ in the main text for readability and because we observed better performance with it. We show how to relate it to the sharpness of the exponential SG in Appendix \ref{app:ivexp}.

\subsection{Applying Condition IV to the exponential SG}
\label{app:ivexp}

We show how we apply equation \ref{appeq:CIV}, to choose the sharpness of an exponential SG. For that we need to define the dependence of the variance of the SG with its sharpness. We use as equivalent notation for the surrogate gradient

\begin{align*}
    \sigma'(v) = \frac{\partial\tilde{H}(v) }{\partial v} = \gamma f(\beta\cdot v)
\end{align*}

We denote no dependency with the voltage in $\sigma'$, when we consider it as a random variable, and we introduce the dependency  $\sigma'(v)$ when we assume the voltage dependence is known. 
The moments of the surrogate gradient are given by

\begin{align}
E[\sigma^{\prime m}]
    =&\int \sigma^{\prime m}p(\sigma') d\sigma'\\
    =&\iint \sigma^{\prime m}p(\sigma'|v)p(v)dv d\sigma' \\
    =&\iint \sigma^{\prime m}\delta\Big(\sigma'-\sigma'(v)\Big) d\sigma'p(v)dv \\
    =&\int \sigma'(v)^mp(v)dv
\end{align}

\noindent where we used the marginalization rule in the second equality and in the third equality we used the fact that $\sigma$ is a deterministic function of $v$, so it inherits its randomness from $v$. We are going to assume as the non-informative prior a uniform distribution between the minimal and maximal values of $y_t-\vartheta$.

\begin{align}
\label{eq:beta}
    E[\sigma'(v)^m]=&\int \sigma'(v)^mp(v)dv\\
    =&\frac{1}{y_{max}-y_{min}}\int^{y_{max}-\vartheta}_{y_{min}-\vartheta} \sigma'(v)^mdv\\
    =&\frac{\gamma^m}{\beta(y_{max}-y_{min})}\int^{\beta(y_{max}-\vartheta)}_{\beta(y_{min}-\vartheta)} f(v')^mdv' 
\end{align}

\noindent where we used the non informative uniform prior assumption in the second equality and we used $\sigma'=\gamma f(\beta v)$ followed by the change of variable $v'=\beta v$ in the third equality. Considering the exponential SG we have that, calling $v_i$ one of the integration limits above, if $v_i$ is positive

\begin{align}
    \int^{v_i}_0 g(|v|)^mdv=&\int^{v_i}_0 g(v)^mdv
\end{align}

\noindent and if $v_i$ is negative

\begin{align}
    \int^{v_i}_0 g(|v|)^mdv=&\int^{v_i}_0 g(-v)^mdv \\
    =&-\int^{-v_i}_0 g(v)^mdv\\
    =&-\int^{|v_i|}_0 g(v)^mdv
\end{align}

\noindent where we made the change of variable $v\rightarrow -v$ in the second equality. Therefore

\begin{align}
    \int^{v_+}_{v_-} g(|v|)^mdv=&sign(v_+)\int^{|v_+|}_0 g(v)^mdv \\
    &- sign(v_-)\int^{|v_-|}_0 g(v)^mdv 
\end{align}

Given that for $v_i>0$ we have

\begin{align}
    \int^{v_i}_0 \text{exponential}(v)^mdv=&\int^{v_i}_0 e^{-2m|v|}dv\\
    =&\int^{v_i}_0 e^{-2mv}dv\\
    =&-\frac{1}{2m}e^{-2mv_i}+\frac{1}{2m}\\
    =&-\frac{1}{2m}e^{-2m|v_i|}+\frac{1}{2m}
\end{align}

\noindent then, for $v_+>0$ and $v_-<0$

\begin{align}\label{eqIVexp}
    \int^{v_+}_{v_-} \text{exponential}(v)^mdv=& \nonumber\\-\frac{1}{2m}e^{-2m|v_+|}-&\frac{1}{2m}e^{-2m|v_-|}+\frac{2}{2m}
\end{align}

\noindent where $v_+=\beta(y_{max}-\vartheta)$ and $v_-=\beta(y_{min}-\vartheta)$
and we show how to compute $y_{max}=Max[y_t]$ and $y_{min}=Min[y_t]$ in section \ref{sec:maxminy}. 
Notice how equation \ref{eq:beta}, shows a dependence of the SG variance  proportional to the square of the dampening and inversely proportional to the sharpness, which recalls our numerical results, where a high sharpness and a low dampening were preferred. Since the dependence with $\beta$ is quite complex, we find the $\beta$ that satisfies the last equation and equation \ref{appeq:CIV} through gradient descent.  This is how condition (IV) is used to fix the sharpness of the exponential SG.

\subsection{Applying Condition IV to the q-PseudoSpike SG}

Instead, when using (IV) to determine the tail-fatness of the SG, we set $\beta=1$ and use 

\begin{align}
    \int^{v_+}_{v_-} &\text{q-PseudoSpike}(|v|)^2dv=\\
    =&\int^{v_+}_{v_-} \text{q-PseudoSpike}(|v|)^2dv\\
    =& \int^{|v_+|}_0 \text{q-PseudoSpike}(v)^2dv \\
    &+ \int^{|v_-|}_0 \text{q-PseudoSpike}(v)^2dv\\
    =&- \frac{q+2|v_+|-1}{2(2q-1)}\frac{1}{\Big(1+\frac{2}{q-1}|v_+|\Big)^{2q}}\nonumber\\
    &- \frac{q+2|v_-|-1}{2(2q-1)}\frac{1}{\Big(1+\frac{2}{q-1}|v_-|\Big)^{2q}}+\frac{q-1}{(2q-1)} \label{eq:IVtail}
\end{align}

When inserted in equation \ref{appeq:CIV}, we use gradient descent to optimize $q$ and find the value that satisfies (IV).

\subsection{Maximal and Minimal voltage values achievable by the network at initialization}
\label{sec:maxminy}

We calculate the maximum and minimum value that the voltage $y$ can take, to be able to complete the argument for condition (IV), about the variance of the backward pass in section \ref{app:ivexp}.
First, we use $Max$ and $Min$ in a statistical ensemble sense, as the maximum/minimum value that a variable could take if sampled over and over again

\begin{align}
    Max[X] =& \sup_{x\sim p(x)} x \\
    Min[X] =& \inf_{x\sim p(x)} x 
\end{align}

When applied to the definition of LIF

\begin{align}
    Max[y_t] =& Max[\alpha_{decay} y_{t-1}(1-x_{t-1})] + Max[W_{rec}x_{t-1}] \nonumber \\&+ Max[b] + Max[W_{in}z_t]  \\
    =& Max[\alpha_{decay}]Max[ y_{t-1}] + (n_{rec}-1)Max[w_{rec}] \nonumber \\&+ Max[b] + n_{in}Max[w_{in}] \label{eq:allfire}\\
    Max[y_t] =& \frac{1}{1-Max[\alpha_{decay}]}\Big((n_{rec}-1)Max[w_{rec}] \nonumber \\&+ Max[b] + n_{in}Max[w_{in}]\Big) \label{eq:nonefire}
\end{align}

\noindent where we used the fact that if $x_t, z_t$ were sampled over and over, the maximum value that they could take is all neurons having fired at the same time, 
we used the fact that $\alpha_{decay}, \vartheta>0$, and we assumed that the maximum is going to stay constant through time $Max[y_{t-1}]=Max[y_{t}]$. Notice that the maximal voltage is achieved when all neurons in the layer fired at $t-1$, equation \ref{eq:allfire}, except for the neuron under study, that stayed silent at $t-1$, to have \ref{eq:nonefire}. Similarly for the bound to the minimal voltage:

\begin{align}
    Min[y_t] =& Min[\alpha_{decay} y_{t-1}(1-x_{t-1})] \nonumber \\&+ (n_{rec}-1)Min[w_{rec}x_{t-1}] + Min[b] \\
    &+ n_{in} Min[w_{in}z_t] \\
    =& Max[\alpha_{decay}]Min[ y_{t-1}] \nonumber \\&+ (n_{rec}-1)Min[w_{rec}] + Min[b] \\&+ n_{in}Min[w_{in}]\\
    Min[y_t] =& \frac{1}{1-Max[\alpha_{decay}]}\Big( (n_{rec}-1)Min[w_{rec}] \nonumber \\&+ Min[b]  + n_{in}Min[w_{in}]\Big) 
\end{align}

Second, we consider another definition of $Max$ and $Min$, where we consider the maximum value achievable by the current sample from the weight distribution. The real maximum value of the voltage will be achieved when the presynaptic neurons to fire are those that are connected with positive weight, we then have that our equation turns to

\begin{align}
    Max[y_t]_i =& \alpha_{decay, i} Max[y_{t-1}]_i + \sum_j ReLU[W_{rec}]_{ij} \nonumber \\&+ b_i + \sum_j ReLU[W_{in}]_{ij} \\
    Max[y_t]_i =& \frac{1}{1-\alpha_{decay,i}}\Big(\sum_j ReLU[W_{rec}]_{ij} \nonumber \\&+ b_i + \sum_j ReLU[W_{in}]_{ij}\Big) 
\end{align}

\noindent where we refer as $\sum_{j} ReLU[W_{rec}]_{ij}$ the sum over columns, where we have typically ommitted the index $i$ for the element of the vector for cleanliness in the rest of the article. The case for the minimum is analogous

\begin{align}
    Min[y_t]_i =& \alpha_{decay,i} Min[y_{t-1}]_i + Min[W_{rec}x_{t-1}] \nonumber \\&+ b_i + Min[W_{in}z_t] \\
    Min[y_t]_i =& \frac{1}{1-\alpha_{decay,i}}\Big( -\sum_{j} ReLU[-W_{rec}]_{ij} \nonumber \\&+ b_i -\sum_{j} ReLU[-W_{in}]_{ij} \Big) 
\end{align}

With this we showed how we calculated the maximal and minimal value of the voltage, to be able to use condition (IV) to define the sharpness of the SG in section \ref{app:ivexp}.

\section{Applying conditions I-IV to an alternative definition of reset}
\label{app:mulreset}

We want to show how the constraints on the weights initialization and on the SG choice change, when the neuron model definition changes. We will use the notation $i_t  = W_{rec}x_t + W_{in}z_t + b$. The reset used by \cite{zenke2021remarkable} is multiplicative to the voltage after having summed the current

\begin{align}
    y_t =&  (\alpha_{decay}y_{t-1} +i_t)(1 - x_{t-1})
\end{align}

\noindent that we will call \textit{post-reset}. Instead,  \cite{wozniak2020deep} uses a LIF with a different definition of reset

\begin{align}
    y_t =&  \alpha_{decay}y_{t-1}(1 - x_{t-1}) +i_t
\end{align}

\noindent that we will call \textit{pre-reset}, since it resets before applying the new current. Another example is given by \cite{lsnn}, that uses a subtractive reset

\begin{align}
    y_t =&  \alpha_{decay}y_{t-1} +i_t -\vartheta x_{t-1} 
\end{align}

\noindent and we will call it \textit{minus-reset}.

The first definition performs as well one refractory period, while the second does not result in a $y_t$ clamped to zero when $x_t=1$.
The factor $(1 - x_t)$ takes the voltage exactly to zero every time the neuron has fired, zero being the equilibrium voltage. What is interesting about this form of reset is that the voltage is reset exactly to $y=0$ after firing, while with the subtractive reset it is not the case. 
We consider training without passing the gradient through the reset, since \cite{zenke2021remarkable} finds better performance in that setting, and it makes the maths cleaner.
The equations that result from the 4 desiderata for this three LIF definitions are as follows

\vspace{.7cm}
\textbf{Post-reset:}
\vspace{-.8cm}

{\small
\begin{align*}
&\hspace{4cm} y_t =  (\alpha_{decay}y_{t-1} +i_t)(1 - x_{t-1})\\  \nonumber \\
    &\overline{w}_{rec}=
 \frac{2}{n_{rec}-1}\Big(1-\alpha_{decay}\Big)\vartheta && \text{I}\\
    &Var[w_{rec}]  =  2(Var[z_t] + \overline{z_t}^2)\frac{n_{in}}{n_{rec}-1}Var[w_{in}] - \frac{1}{2}\overline{w}_{rec}^2 && \text{II}\\
    &\gamma=   \frac{1}{(n_{rec}-1)\hat{w}_{rec}}\Big( 1 -\alpha_{decay}-\xi n_{in}\hat{w}_{in} \gamma_{in} \Big)&& \text{III}\\
    &\overline{\sigma^{\prime 2}} = \frac{2-\alpha_{decay}^2-\xi n_{in}\overline{w_{in}^2} \ \overline{\sigma^{\prime 2}_{in}}}{(n_{rec}-1)\overline{w^2}_{rec}}  && \text{IV}
\end{align*}
}

\vspace{.7cm}
\textbf{Pre-reset:}
\vspace{-.8cm}

{\small
\begin{align*}
&\hspace{4cm} y_t =  \alpha_{decay}y_{t-1}(1 - x_{t-1}) +i_t\\  \nonumber \\
    &\overline{w}_{rec}=
    \frac{1}{n_{rec}-1}\Big(2-\alpha_{decay}\Big)\vartheta && \text{I}\\
    &Var[w_{rec}]  =  2(Var[z_t] + \overline{z}_t^2)\frac{n_{in}}{n_{rec}-1}Var[w_{in}] - \frac{1}{2}\overline{w}_{rec}^2 && \text{II}\\
    &\gamma=   \frac{1}{(n_{rec}-1)\hat{w}_{rec}}\Big( 1 -\alpha_{decay}-\xi n_{in}\hat{w}_{in} \gamma_{in} \Big)&& \text{III}\\
    &\overline{\sigma^{\prime 2}} = \frac{1-\frac{1}{2}\alpha_{decay}^2-\xi n_{in}\overline{w_{in}^2} \ \overline{\sigma^{\prime 2}_{in}}}{(n_{rec}-1)\overline{w^2}_{rec}}  && \text{IV}
\end{align*}
}

\vspace{.7cm}
\textbf{Minus-reset:}
\vspace{-.8cm}

{\small
\begin{align*}
&\hspace{4cm} y_t = \ \alpha_{decay}y_{t-1} +i_t -\vartheta x_{t-1} \\ \nonumber \\
    &\overline{w}_{rec} = 
    \frac{1}{n_{rec}-1}\Big(3-2\alpha_{decay}\Big) \vartheta && \text{I}\\
    &Var[w_{rec}]  =  2(Var[z_t] + \overline{z}_t^2)\frac{n_{in}}{n_{rec}-1}Var[w_{in}] - \frac{1}{2}\overline{w}_{rec}^2 && \text{II}\\
    &\gamma =    \frac{1}{(n_{rec}-1)\check{w}_{rec}-\vartheta}\frac{\check{w}_{rec}}{\hat{w}_{rec}}\Big( 1 -\alpha_{decay}-\xi n_{in}\hat{w}_{in} \gamma_{in} \Big)&& \text{III}\\
    &\overline{\sigma^{\prime2}} =   \frac{1-\alpha_{decay}^2-\xi n_{in}\overline{w_{in}^2}\ \overline{\sigma_{in}^{\prime2}}}{(n_{rec}-1)\overline{w^2}_{rec} + \vartheta^2}  && \text{IV}
\end{align*}
}

To have the conditions when the gradient does not pass through the reset, put $\vartheta=0$ in (III) and (IV), but not in (I).

\section{ALIF and sLSTM models}
\label{app:alifsLSTM}

To study the variability of SG training with the architecture of choice, we tested different SG shapes on the ALIF and sLSTM networks.
We used the following ALIF definition

\begin{align}    
\boldsymbol{y}_{t,l} =& \boldsymbol{\alpha}_{decay,l}^y \boldsymbol{y}_{t-1,l}  \nonumber\\ &+W_{rec,l}\boldsymbol{x}_{t-1,l}  + W_{in,l}\boldsymbol{x}_{t-1,l-1} + \boldsymbol{b}_l \nonumber\\&- \boldsymbol{\vartheta}_{t-1,l} \ \boldsymbol{x}_{t-1,l}\\
\boldsymbol{\vartheta}_{t,l} =& \boldsymbol{\alpha}^{\vartheta}_{decay,l} \boldsymbol{\vartheta}_{t-1,l} +\boldsymbol{b}^{\vartheta}_l + \boldsymbol{\beta}_l \boldsymbol{x}_{t-1,l} 
\end{align}    

\vspace{.5cm}

\noindent where we initialized $W_{rec}, W_{in}$ as Glorot Uniform, $b_l=0$, $\alpha_{decay,l}^y=4\cdot10^{-5}$,  $\alpha_{decay,l}^\vartheta=0.992$ for the SHD task and $\alpha_{decay,l}^\vartheta=0.98$ for the sl-MNIST task, $b^{\vartheta}_l =0.01$, and $\beta_l=1.8$.

The LSTM implementation that we used is the following

\begin{align}
    i_t =& \sigma_g(W_ix_t + U_ih_{t-1}+b_i)\\
    f_t =& \sigma_g(W_fx_t + U_fh_{t-1}+b_f)\\
    o_t =& \sigma_g(W_ox_t + U_oh_{t-1}+b_o)\\
    \tilde{c}_t =& \sigma_c(W_cx_t + U_ch_{t-1}+b_c)\\
    c_t =& f_t\circ c_{t-1}+ i_t\circ \tilde{c}_t\\
    h_t =& o_t\circ\sigma_h(c_t)
\end{align}

\vspace{.5cm}

The dynamical variables $i_t, f_t, o_t$ represent the input, forget and output gates, that prevent representations and gradients from exploding, while $c_t, h_t$ represent the two hidden layers of the LSTM, that work as the working memory and are maintained and updated through data time $t$.
To construct the spiking version of the LSTM (sLSTM) we turned  the activations into $\sigma_g(x)=H(x)$ and $\sigma_c=\sigma_h=2H(x)-1$. The matrices $W_j, U_j$ are initialized with Glorot Uniform initialization, and the biases $b_j$ as zeros, with $j\in\{i, f, o, c\}$.

\section{More on Sparsity}\label{app:more_sparsity}

\begin{figure}
    \centering
    \vspace{.55cm}
    \includegraphics[width=.45\textwidth]{images/surrogate_grads/t__sparsity_tsgfastsigmoidpseudod_tSHD.pdf}
    \includegraphics[width=.45\textwidth]{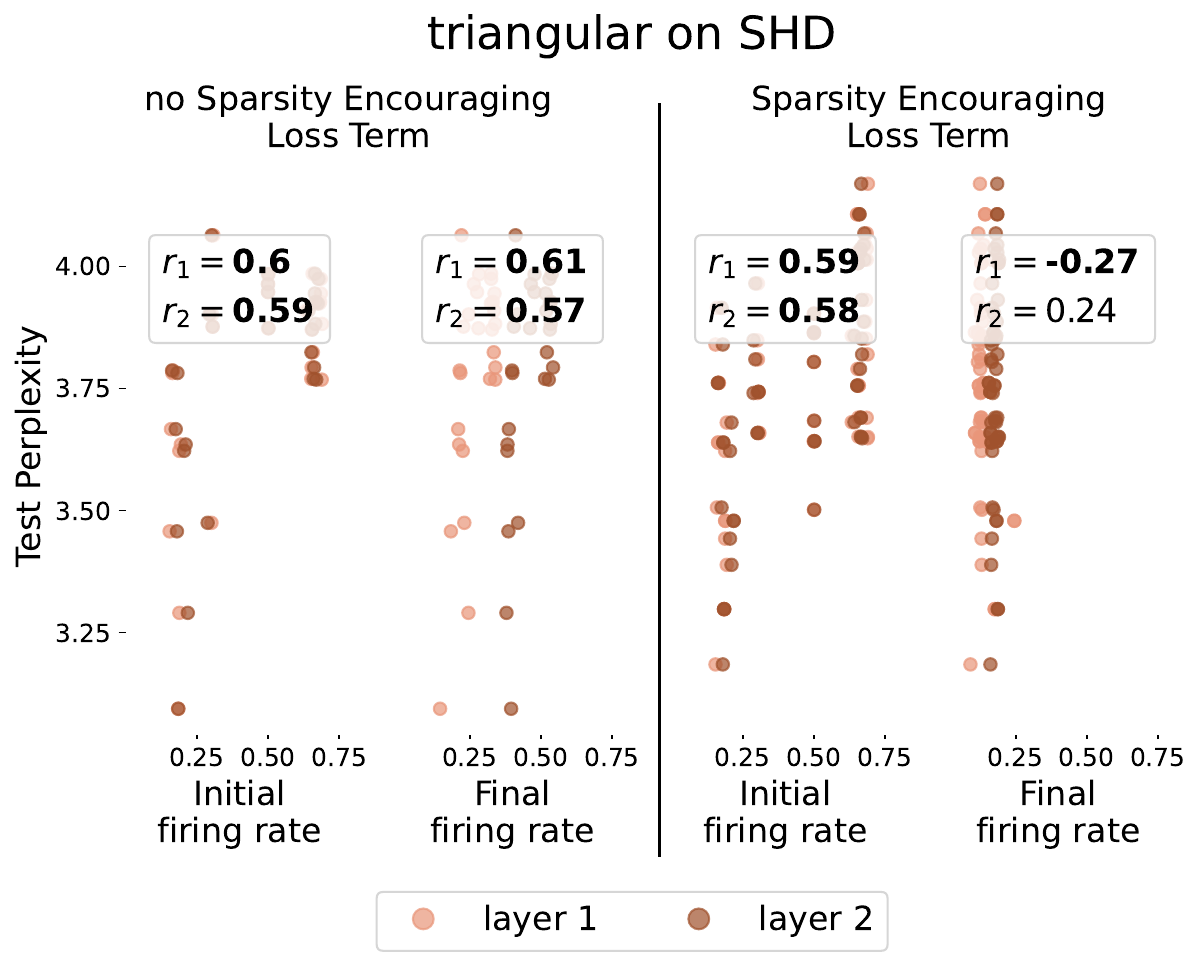}
    \includegraphics[width=.45\textwidth]{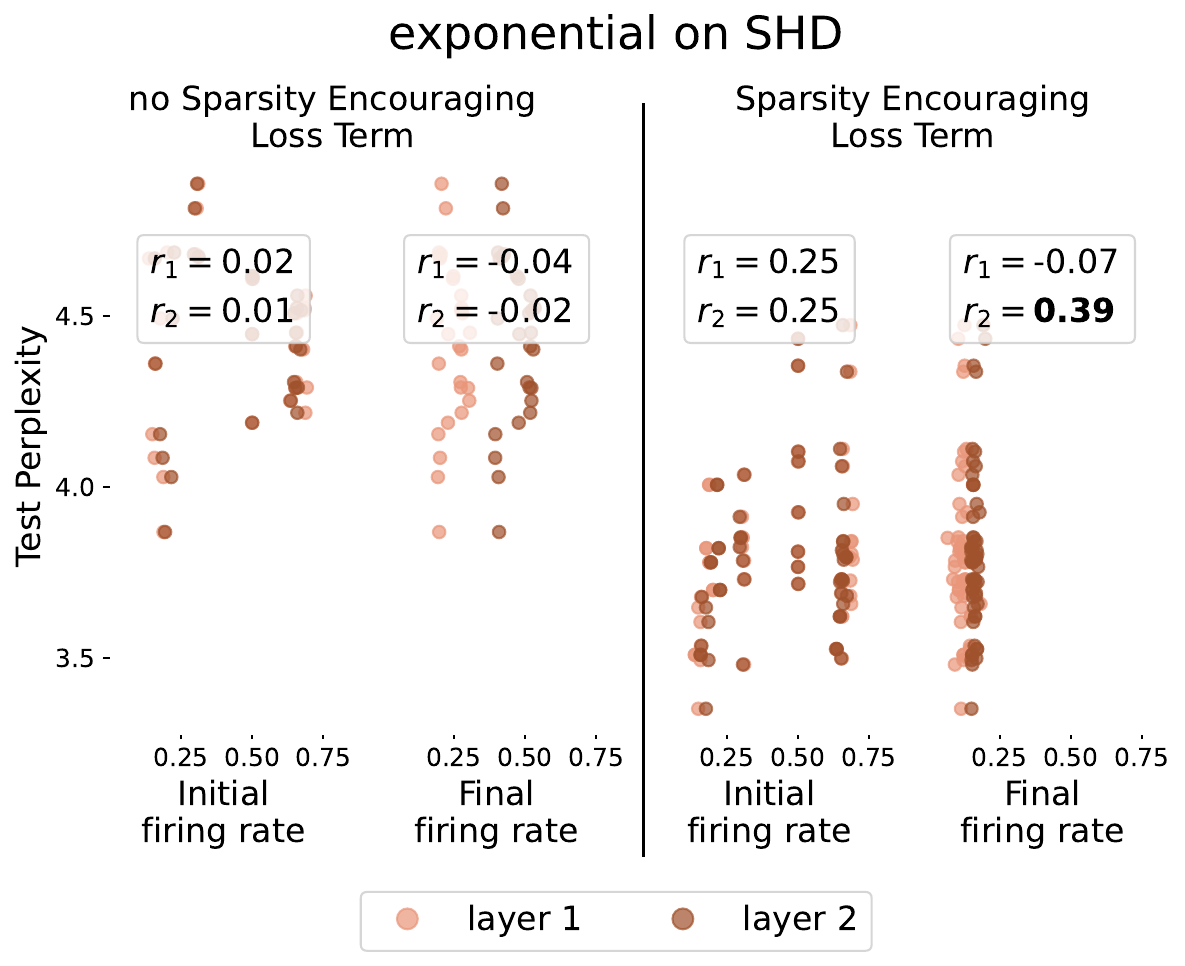}
    \caption{\textbf{Sparsity role is not consistent across SG shapes.} When we fix the task to be the SHD task, we see that the preference for high or low firing rates at initialization and after training on the test set, depends on the SG of choice. As we saw in the main text, 
    the derivative of the fast sigmoid has preference for high $\rho_i$, since, at each layer $l$, the final loss correlation $r_l$ with the firing rate at initialization is negative.
    Instead, the triangular SG has preference for low $\rho_i$ since the correlation is positive, while for the exponential SG, $\rho_i$ does not seem to correlate with final performance, given that the correlations $r_l$ are not significant (in bold when significant).}
    \label{fig:sparsity_and_shape}
\end{figure}

\begin{figure}
    \centering
    \includegraphics[width=.45\textwidth]{images/surrogate_grads/t__sparsity_tsgfastsigmoidpseudod_tSHD.pdf}
    \includegraphics[width=.45\textwidth]{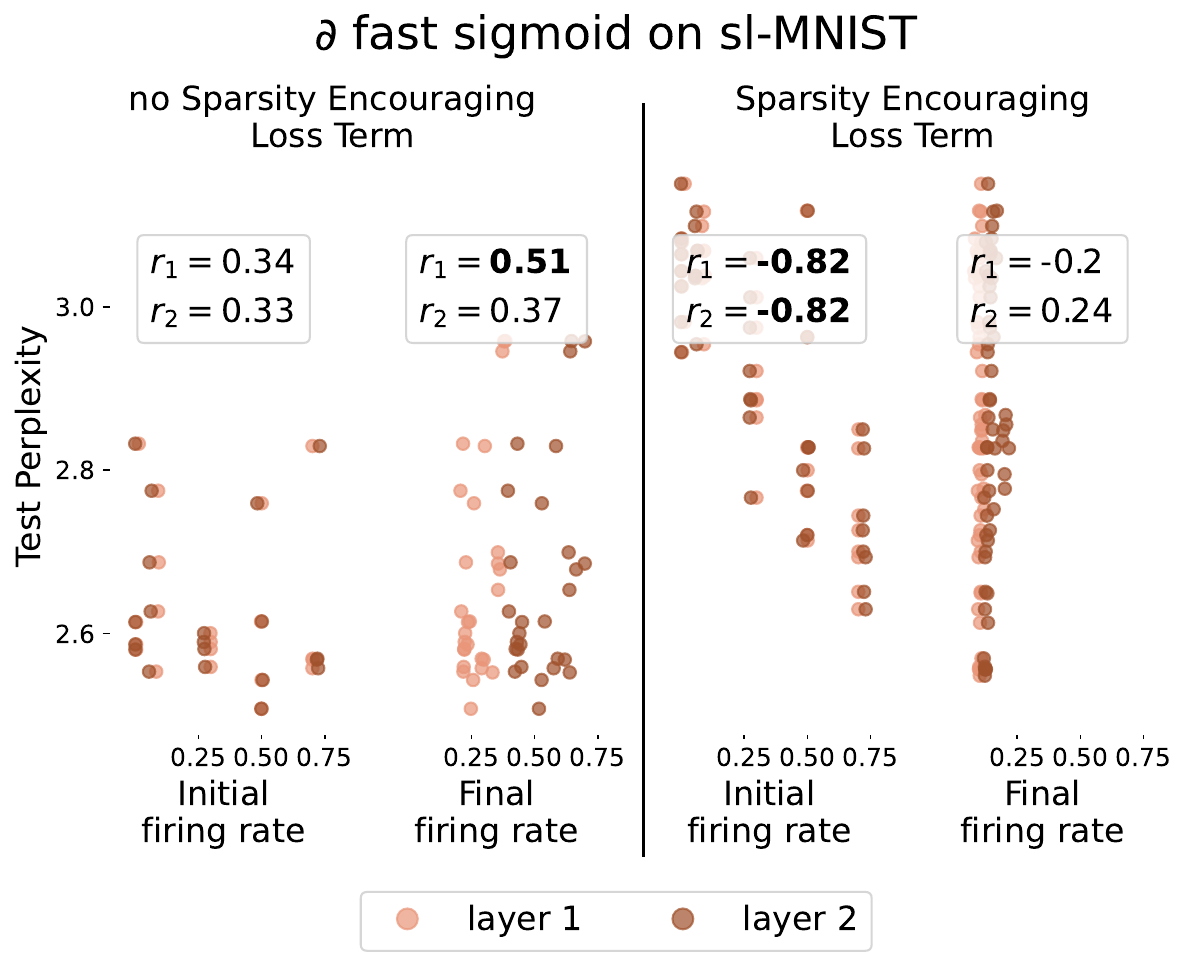}
    \includegraphics[width=.45\textwidth]{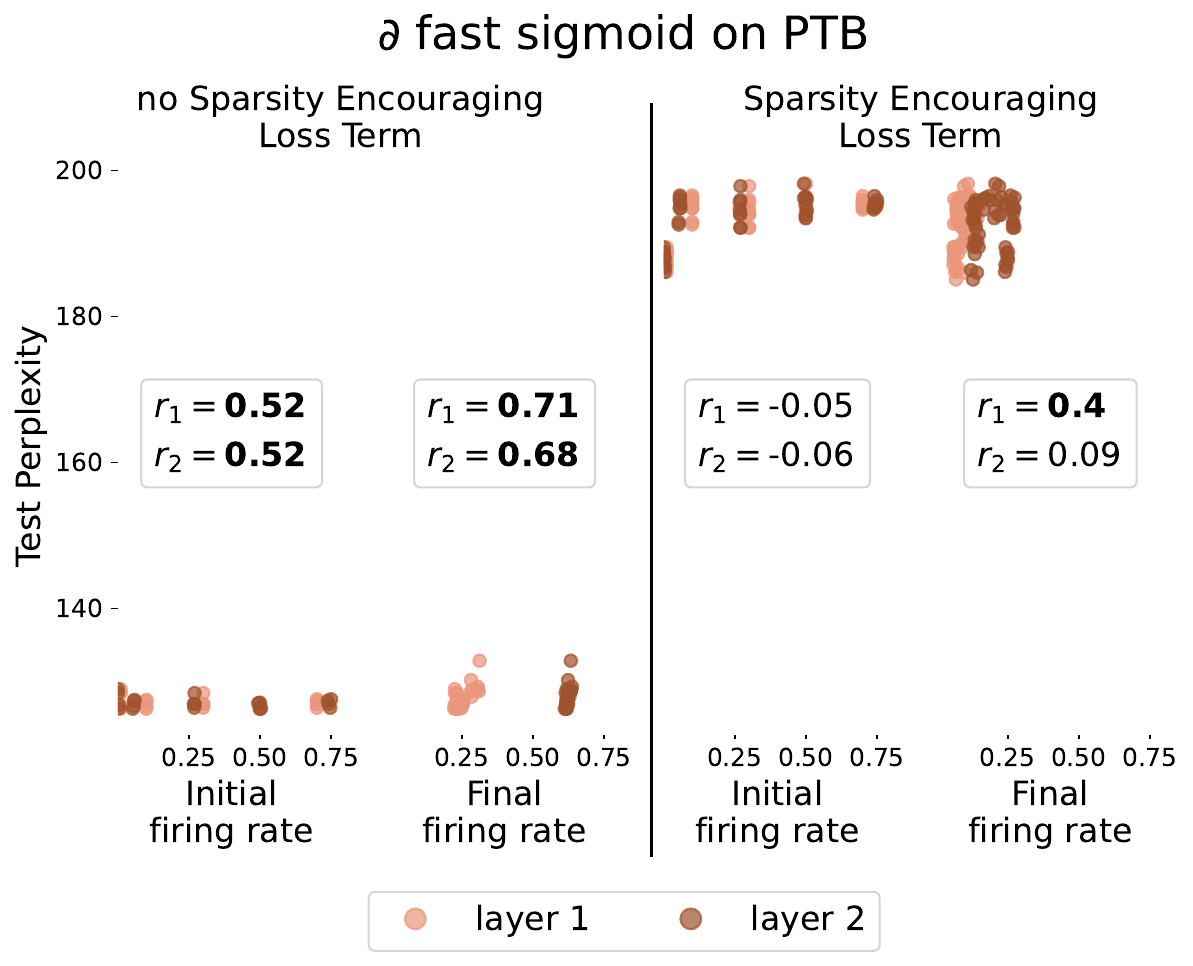}
    \caption{\textbf{Sparsity role is consistent across tasks.} Here we fix the SG shape to the derivative of the fast sigmoid and we change the task. On sl-MNIST, we see a similar trend than on SHD, where high initial firing rate is preferred for better performance when sparsity is encouraged. Encouraging sparsity has a negative effect on learning language modeling on the PTB task. However, when no sparsity is encouraged, best performance on PTB is still at $\rho_i=0.5$.}
    \label{fig:sparsity_and_task}
\end{figure}

We investigate if the role of sparsity remains consistent across SG shapes in Fig. \ref{fig:sparsity_and_shape}, and across tasks in Fig. \ref{fig:sparsity_and_task}. Notice that Fig. \ref{fig:sparsity} is repeated in Fig. \ref{fig:sparsity_and_shape} and \ref{fig:sparsity_and_task} to ease the comparison.

\end{document}